\documentclass[lettersize,journal]{IEEEtran}
\usepackage{amsmath,amsfonts}
\usepackage{algorithmic}
\usepackage{algorithm}
\usepackage{array}
\usepackage[caption=false,font=normalsize,labelfont=sf,textfont=sf]{subfig}
\usepackage{textcomp}
\usepackage{stfloats}
\usepackage{url}
\usepackage{verbatim}
\usepackage{graphicx}
\usepackage{cite}
\usepackage{eccvabbrv}
\usepackage{multirow}
\usepackage{color}
\begin{document}

\title{Skull-to-Face: Anatomy-Guided 3D Facial Reconstruction and Editing}
\author{
    Yongqing Liang,
    Congyi Zhang,
    Junli Zhao,
    Wenping Wang,
    Xin Li$^{*}$\thanks{$^*$Corresponding author.}
\thanks{Yongqing Liang, Wenping Wang are from the Department of Computer Science and Engineering at Texas A\&M University. Emails: \{lyq, wenping\}@tamu.edu. Congyi Zhang is from the Department of Computer Science at the University of British Columbia. Email: congyiz@cs.ubc.ca. Junli Zhao is from the College of Computer Science and Technology at Qingdao University. Email: zhaojl@yeah.net. Xin Li is from the College of Performance, Visualization, \& Fine Arts at Texas A\&M University. Email: xinli@tamu.edu.}
\thanks{}
}

\markboth{Journal of \LaTeX\ Class Files,~Vol.~14, No.~8, August~2021}%
{Liang \MakeLowercase{\textit{et al.}}: Skull-to-Face: Anatomy-Guided 3D Facial Reconstruction and Editing}


\maketitle

\newcommand{\edit}[1]{{#1}}

\begin{abstract}

Deducing the 3D face from a skull is a challenging task in forensic science and archaeology. 
This paper proposes an end-to-end 3D face reconstruction pipeline and an exploration method that can conveniently create textured, realistic faces that match the given skull.
To this end, we propose a tissue-guided face creation and adaptation scheme.   
With the help of the state-of-the-art text-to-image diffusion model and parametric face model, we first generate an initial reference 3D face, whose biological profile aligns with the given skull. 
Then, with the help of tissue thickness distribution, we modify these initial faces to match the skull through a latent optimization process. 
The joint distribution of tissue thickness is learned on a set of skull landmarks using a collection of scanned skull-face pairs. 
We also develop an efficient face adaptation tool to allow users to interactively adjust tissue thickness either globally or at local regions to explore different plausible faces. 
Experiments conducted on a real skull-face dataset demonstrated the effectiveness of our proposed pipeline in terms of reconstruction accuracy, diversity, and stability. 
Our project page is \url{https://xmlyqing00.github.io/skull-to-face-page/}.

\end{abstract}

\section{Introduction}
\label{sec:introduction}

\begin{figure}
    \centering
    \includegraphics[width=\linewidth]{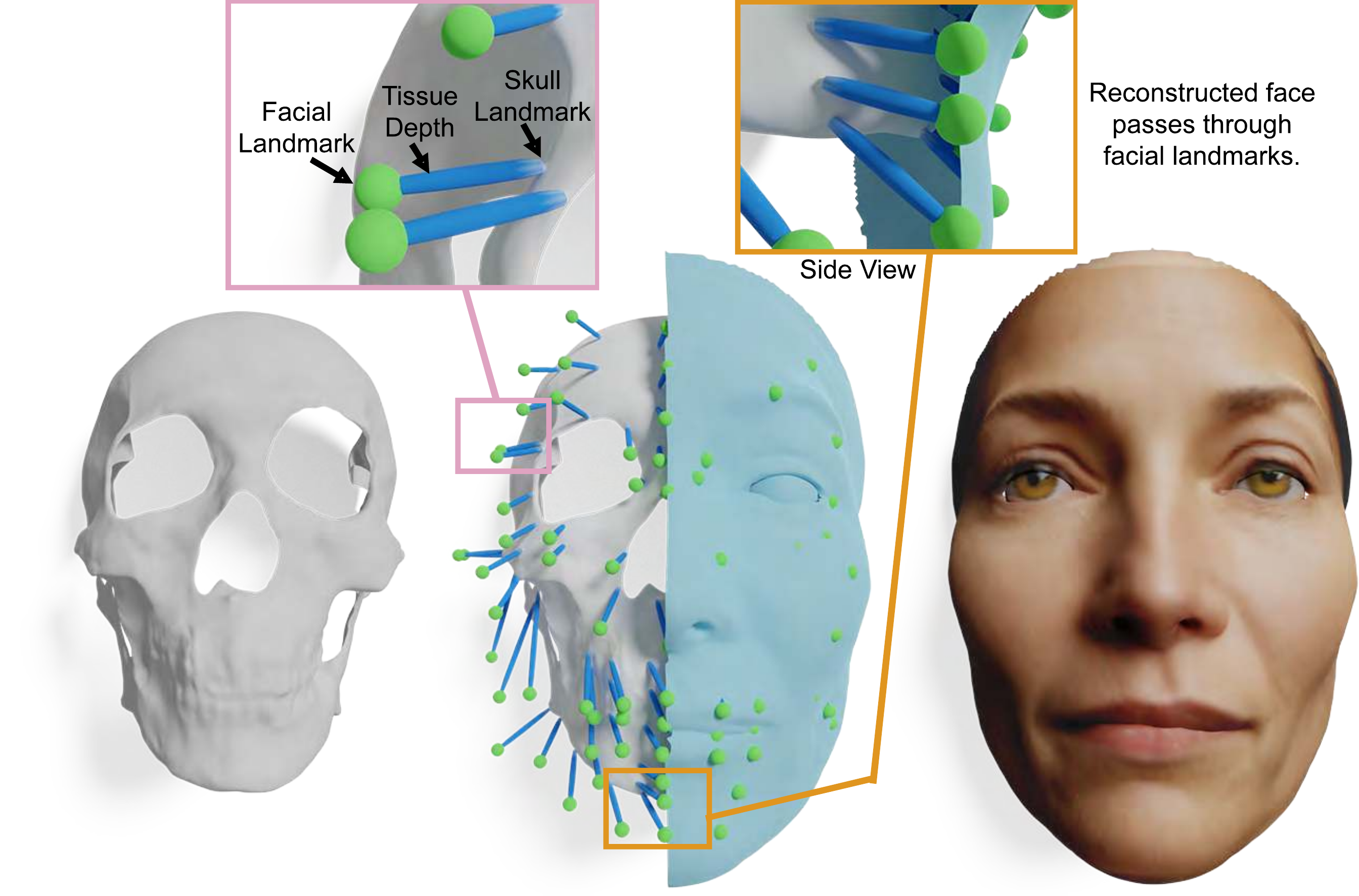}
    \caption{Given an input skull scan, based on skull landmarks and tissue depth (blue sticks), facial landmarks (green spheres) can be obtained. The 3D face is inferred under the constraints of facial landmarks.}
    \label{fig:demo}
\end{figure}

3D facial reconstruction from a skull is an effective approach used by forensic investigators to identify skeletal remains when other information is unavailable. 
It has been successfully applied in forensic cases~\cite{vandermeulen2012automated} and anthropological studies~\cite{aidonis2023digital}. 
Given a 3D skull scan, the objective is to derive a 3D face that aligns with both the geometric features and the biological profile of the skull. 
The anatomic setting of the reconstruction process is illustrated in Fig.~\ref{fig:demo}: 
A set of predefined \emph{skull landmarks}~\cite{Deng11FSI} is distributed on the query \emph{skull}. 
The \emph{tissue depths} (blue sticks) on landmarks are estimated using anatomical statistic data. 
The endpoints of these tissue depths, \ie, the \emph{facial landmarks} (green spheres), depict a coarse geometry shape of the 3D face. 
A plausible 3D face geometry should be created, satisfying the constraints of these facial landmarks.

Traditional facial reconstruction in forensic applications is performed manually using plasticine or clay~\cite{verze2009history, joukal2015facial}.
Recent approaches~\cite{baldasso20213d, miranda2018assessment, wilkinson2019search} employ a digital environment
to replace these manual clay modeling methods. 
Both physical and digital methods for facial reconstruction require years of anatomy and artistic modeling experience, and the results are often unreliable due to the subjective factors~\cite{gupta2015forensic, zhang2022end}.
Besides the geometry, a vivid facial texture may be inferred from (\eg, ancestry, age of) the skull. 
Such realistic facial textures can be desirable in anthropological and forensic applications, but their creation often requires further post-processing steps and significant extra effort.

In the past two decades, researchers have been studying computer-aided skull-to-face reconstruction algorithms to produce 3D faces.  
They access the problem from various perspectives and multidisciplinary backgrounds, from data-driven~\cite{Face:Claes06FSI, Face:PaysanDAGM09, cao2014facewarehouse, deng_regional_2016, gietzen2019method, zhang2022end} to physical simulation~\cite{Face:Jones01VMVC, Face:KahlerSIG03, Face:Muller05ISPA, Deng11FSI}, providing solid research baselines. 
Nevertheless, (1) all these face generation methods overlooked multi-modal characteristics beyond the geometry of the skull, such as gender, age, and ancestry. 
(2) Without modeling the distribution of tissue depth, these methods addressed the problem by solving a deterministic mapping from one skull to one face with statistically average tissue depths \cite{rhine1980thickness, rhine2001tables,stephan2015accuracies,stephan2016facial}, making it hard for further tuning in real scenarios. 

Our proposed work builds upon two main observations. 
(1) Firstly, the biological profile, encompassing ancestry, age, gender, and stature of the body, can be estimated from skulls~\cite{spradley2016metric}. 
Different population groups exhibit distinct facial priors, and we argue that incorporating these is crucial for accurate facial reconstruction.
We generate an initial guess of the 3D face with these attributes, whereas existing methods often overlook this key information. 
(2) Secondly, the combinations of tissue depths in real human faces follow intrinsic distributions that have not been well-studied. 
We proposed a 3D face representation based on the skull and the tissue depth distribution models.
This representation can precisely control the geometry shape on the 3D face.
We introduce an additional control based on the distribution of tissue depth over facial reconstruction, providing an intuitive and effective tool for 3D face manipulation and exploration.

Our approach includes an offline training stage for tissue depth distribution and two-stage online inference for face reconstruction: semantic face model initialization, and anatomy-guided face adaptation. We also develop landmark-based face editing as an optional adjustment stage.
(1) During the preprocessing stage, tissue depth distribution can be analyzed using available datasets of scanned skull and face pairs, or obtained from statistical data in forensic anthropology literature~\cite{rhine1980thickness, rhine2001tables,stephan2015accuracies,stephan2016facial,caple2016standardized,thiemann2017vivo}. 
(2) When processing a query skull, the key biological profile of the skull is used to generate a 2D realistic facial image, which is then reconstructed into a textured 3D face. Then, facial landmarks are generated, guided by the pre-trained tissue depth distribution and the geometry of the query skull. Finally, the face geometry is adjusted to align with these facial landmarks, resulting in the final skull-conformed face.
(3) As an optional step, we provide a tool that allows users to intuitively adjust the face geometry by tuning a one-dimensional feature in the distribution space of tissue depths.

The \textbf{main contributions} of this work are as follows: 
\begin{itemize}
\item[(1)] We present a novel face generation and adaptation pipeline that can effectively reconstruct a face from a given skull. 
This pipeline incorporates a new 3D face representation where faces are defined by learned tissue depth (thickness) distributions. 
This pipeline supports 
(a) generation of faces that accurately follow geometry constraint from the skull, 
and (b) intuitive exploration of plausible faces that match this skull. 
\item[(2)] We leverage foundation models to create a bio-profile-aligned initial face. By learning tissue depth distributions, we design an effective face adaption module to modify the initial face to match the query skull. 
\item[(3)] The learned tissue depth distribution also allows us to build a control mechanism that offers an intuitive and effective way to explore different plausible faces.
\item[(4)] Compared to existing methods, our method achieves state-of-the-art geometric accuracy qualitatively and quantitatively. 
We will release the codes and pre-trained models, making it the first publicly accessible approach in this field. 
\end{itemize}

\section{Related Work}
\label{sec:relatedwork}

\subsection{Facial Reconstruction from Skull}

Statistics-based methods for facial reconstruction employ high-dimensional vectors, typically representing faces through their 3D coordinates or feature points. 
These methods, using techniques such as PCA, learn the mapping between corresponding skulls and faces from training datasets~\cite{Face:Claes06FSI, Face:PaysanDAGM09, cao2014facewarehouse, deng_regional_2016, zhao2017craniofacial, madsen2018probabilistic, gietzen2019method, HF-GGR:jia2021craniofacial}. 
A new face is then generated as a linear combination of principal components.
Other approaches~\cite{madsen2018probabilistic, duan20153d, berar2011craniofacial, Face:PaysanDAGM09, tilotta2010local, liu2018superimposition, qiu2022sculptor} involve regression algorithms to predict faces from skulls. 
Zhang~\etal~\cite{zhang2022end} project 3D skulls to 2D images and leverage image-to-image style transferring networks~\cite{isola2017pix2pix, CycleGAN2017} for facial reconstruction. 
\edit{
SCULPTOR~\cite{qiu2022sculptor} learns a joint distribution of skull and face from the annotated dataset, but it's hard to integrate it into our system because only a few sample data and partial codes are publicly available. 
}
However, these data-driven methods require a significant number of joint skull-face scans, which are costly and difficult to acquire. 
The limited available datasets often result in overfitted networks and dataset bias.

Another category of reconstruction algorithms employs 3D modeling techniques to digitally mimic the manual reconstruction process. 
A prevalent strategy involves deforming a template face surface~\cite{Deng11FSI, Face:Muller05ISPA} or tissue volume~\cite{valsecchi2018robust, Face:KahlerSIG03, Face:Jones01VMVC} to fit with the subject's skull. 
This deformation is controlled by imposing penalties on certain geometric smoothness energy to regularize the transformation while adhering to the positional constraints of facial landmarks.

\subsection{\edit{Physics-based Face Modeling}}

\edit{
Recent physics-based approaches for facial animation can be categorized into two types: \emph{jaw-driven} and \emph{muscle-driven}.
Zoss~\etal~\cite{zoss2019accurate} presents a method to learn the non-linear mapping from the skin deformation to the underlying jaw motion on an annotated dataset.
This mapping can be re-targeted to new subjects to guide the facial animation.
Sifakis~\etal~cite{sifakis2005automatic} built a model of facial musculature, passive tissue, and underlying skeletal structure using volumetric data acquired from a living male subject.
They used the finite element model to drive the face animation by muscle activation.
Ichim~\etal~\cite{ichim2017phace} model the physical interaction of passive flesh, active muscles, and rigid bone structures by minimizing a set of non-linear potential energies.
The learned muscle activation model leads to a robust reproduction of complex facial articulations.
Kadle{\v{c}}ek and Kavan~\cite{kadlevcek2019building} built a volumetric model from magnetic resonance images of a neutral facial expression.
They solved an inverse physics problem to learn the mechanical properties of the face from the training data.
Our method provides another solution that uses tissue depths to control the shape of the human face.
}

\subsection{Digital Face Synthesis}

Constrained face editing or re-synthesis techniques involve generating new faces that are both realistic and adhere to specific constraints.
Many recent 2D-based methods~\cite{alaluf2021matter-sam, Karras2021, patashnik2021styleclip, dhariwal2021diffusion, sehwag2022generating, rombach2022high, dagan-hong2022depth} accomplish this by manipulating semantic codes in the latent space to generate new 2D face images. 
However, these 2D methods lack underlying 3D face geometry, making it difficult to enforce fine geometry constraints during face editing. 

Recent years have witnessed significant advancements in large face generation models, including text-to-image generation models (\eg Stable Diffusion XL~\cite{anonymous2023sdxl} and DALL-E2~\cite{ramesh2022hierarchical}), and 2D image to 3D face generation models (\eg, GaGAN~\cite{kossaifi2018gagan}, CoMA~\cite{ranjan2018generating}, DECA~\cite{DECA:feng2021learning}, EG3D~\cite{eg3d:Chan2022}, and PanoHead~\cite{Panohead:An_2023_CVPR}). A recent study,  Rodin~\cite{wang2023rodin}, proposed a text-to-3D-face pipeline, but the synthesized faces are somewhat cartoon-like and unrealistic. 
NeRF-based methods such as EG3D and PanoHead are less effective in editing 3D geometry (to meet specific geometric constraints). 

\section{Method}

\begin{figure*}[t]
\centering
    \includegraphics[width=\linewidth]{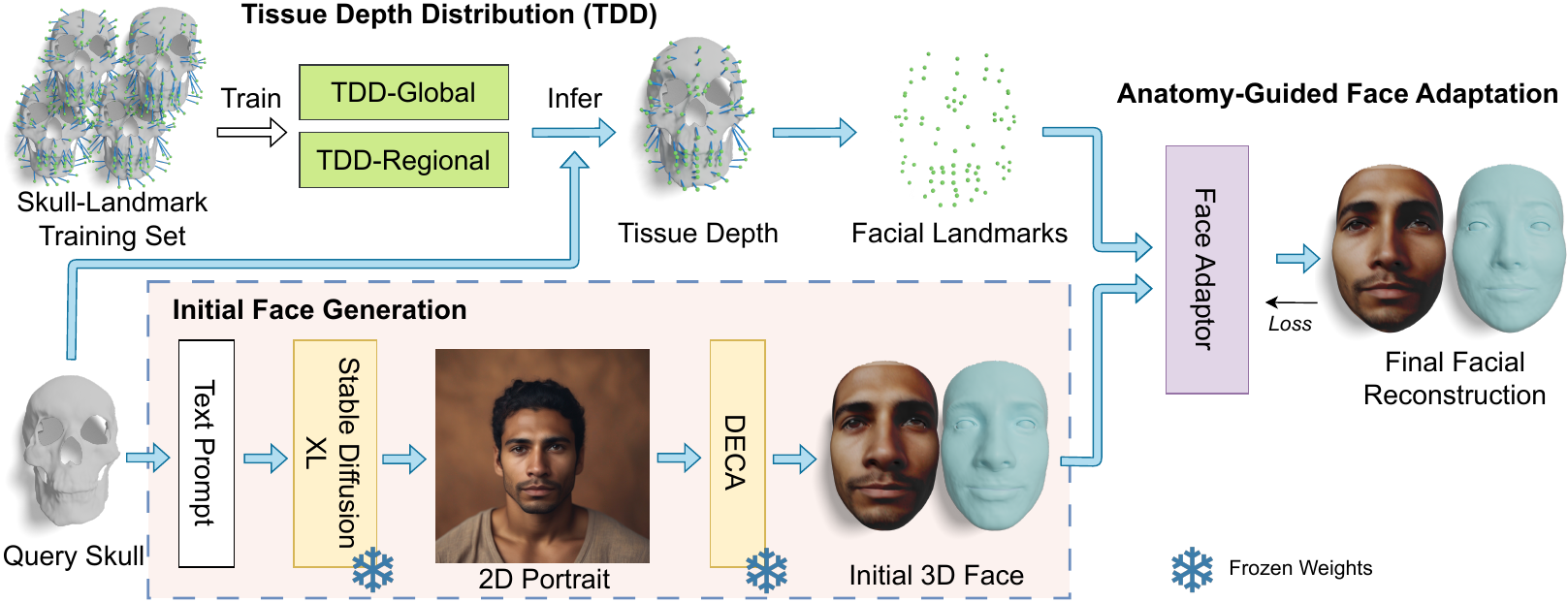}
    \caption{An overview of the proposed Skull-to-Face pipeline, which includes three modules. 
    The Face Generation module synthesizes an initial 3D face in consideration of the biological profile of the skull such as age, gender, and ancestry.
    The Tissue Depth Distribution (TDD) can suggest a valid combination of facial landmarks as geometry constraints.
    The Anatomy-guided Face adaptation module optimizes the initial face as the facial reconstruction according to the landmark constraints.}
\label{fig:overview}
\end{figure*}

\subsection{Pipeline Overview}

Our skull-to-face generation pipeline has three components, as is shown in Fig.\ref{fig:overview}. 
In the \emph{Tissue Depth Distribution} (TDD) modeling, we analyze the tissue depth distribution in the training set and use the distribution to suggest a valid combination of tissue depths for the skull (Sec.~\ref{sec:tissue-depth}).
Then, when reconstructing a face from a query skull, we have two stages. 
In the \emph{Initial Face Generation} stage, we utilize key biological profiles about the human body, \ie, gender, age, and ancestry to generate a 2D candidate portrait using Stable Diffusion XL~\cite{anonymous2023sdxl} and lift it to a 3D face using the FLAME model~\cite{DECA:feng2021learning} (Sec.~\ref{sec:face-generation}).
Finally, with the facial landmarks computed from the tissue depth model as the geometry constraints, we propose an \emph{Anatomy-guided Face Adaptation} method to modify the 3D face according to facial landmark constraints. 
The details of the reconstructed face are preserved from the 2D candidate portrait image (Sec.~\ref{sec:face-editing}).

\subsection{Distribution of Tissue Depth}
\label{sec:tissue-depth}

Tissue depths define the displacement between facial landmarks and their corresponding skull landmarks.
The combination of tissue depths controls the shape of the reconstructed face, which is critical for the fidelity of facial reconstruction and directly influences the reconstruction error. 
Instead of directly learning the mapping from skull to face as other works~\cite{HF-GGR:jia2021craniofacial, zhang2022end} did, 
we analyzed and explored the intrinsic pattern of the tissue depth distribution.
We then used the distribution pattern to guide facial reconstruction.
This also provides the benefits for users to further adjust tissue depths to explore various plausible faces on skulls.

On each skull $J$, a set of landmarks $\mathcal{P}^s=\{ \mathbf{p}^s_i \in \mathbb{R}^2, i=1,\ldots, n\}$ are defined and can be extracted. In this work, we follow \cite{Deng11FSI} to extract $n=78$ landmarks. 
Each such skull landmark $\mathbf{p}^s_i$ is extended outward along the normal directions $\hat{\mathbf{n}}^s_i$ with a tissue depth length $d_i$. 
The extended landmarks are referred to as facial landmarks $\mathcal{P}^f = \{\mathbf{p}^f_i | \mathbf{p}^f_i=\mathbf{p}^s_i + d_i\hat{\mathbf{n}}^s_i\}$, which provide a sparse constraint of the face shape. 
$\mathcal{D} = \{d_i\}$ represents the tissue depths of a person, and 
$\mathcal{X} = \{\mathcal{D}_m\} \in \mathbb{R}^n$ represents the collection of tissue depths of $m$ people in the training dataset.

\edit{
Medical and forensic studies have collected several soft tissue depth databases~\cite{caple2016standardized,thiemann2017vivo}  that serve as valuable references for tissue depth. 
These databases enable inference of facial tissue depth based on factors such as BMI, sex, age, and regional population.
This option allows users to retrieve corresponding tissue depths from the database when available. 
However, since a comprehensive database that includes all population groups is unfeasible, we provide a generalized learning model to represent the typical distribution of human tissue depths.
}

\subsubsection{TDD-Global Model}

We analyze the tissue depth distribution using the training set of the skull-face dataset, which comprises $m$ samples, each with $n=78$ landmarks. 
It is important to note that the number of training samples $m$ in the publicly accessible skull-face dataset is limited ($m \leq 100$ in our experiments). Consequently, employing a categorized or neural-based model to fit the landmark distribution may be inappropriate. 
We therefore designed a simpler model to encode the tissue depth distribution. 

\begin{figure}
    \centering
    \begin{tabular}{c}
         \includegraphics[width=\linewidth]{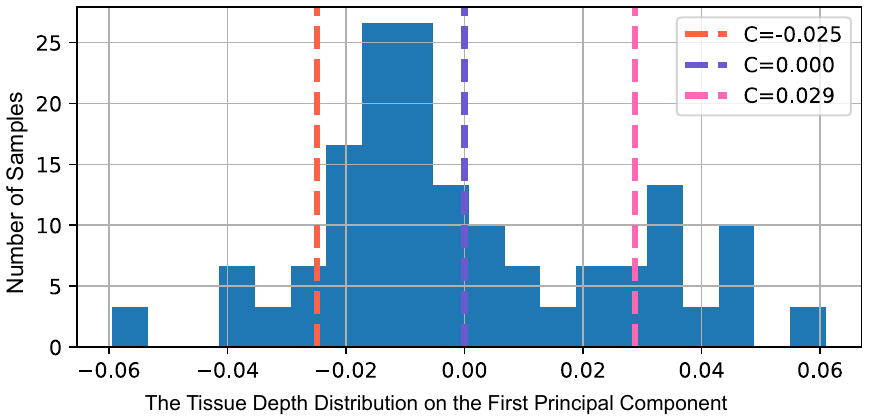}\\
         (a) Three representative values of the first principal component\\are visualized overlaid on the tissue depth distribution.\\
         \includegraphics[width=\linewidth]{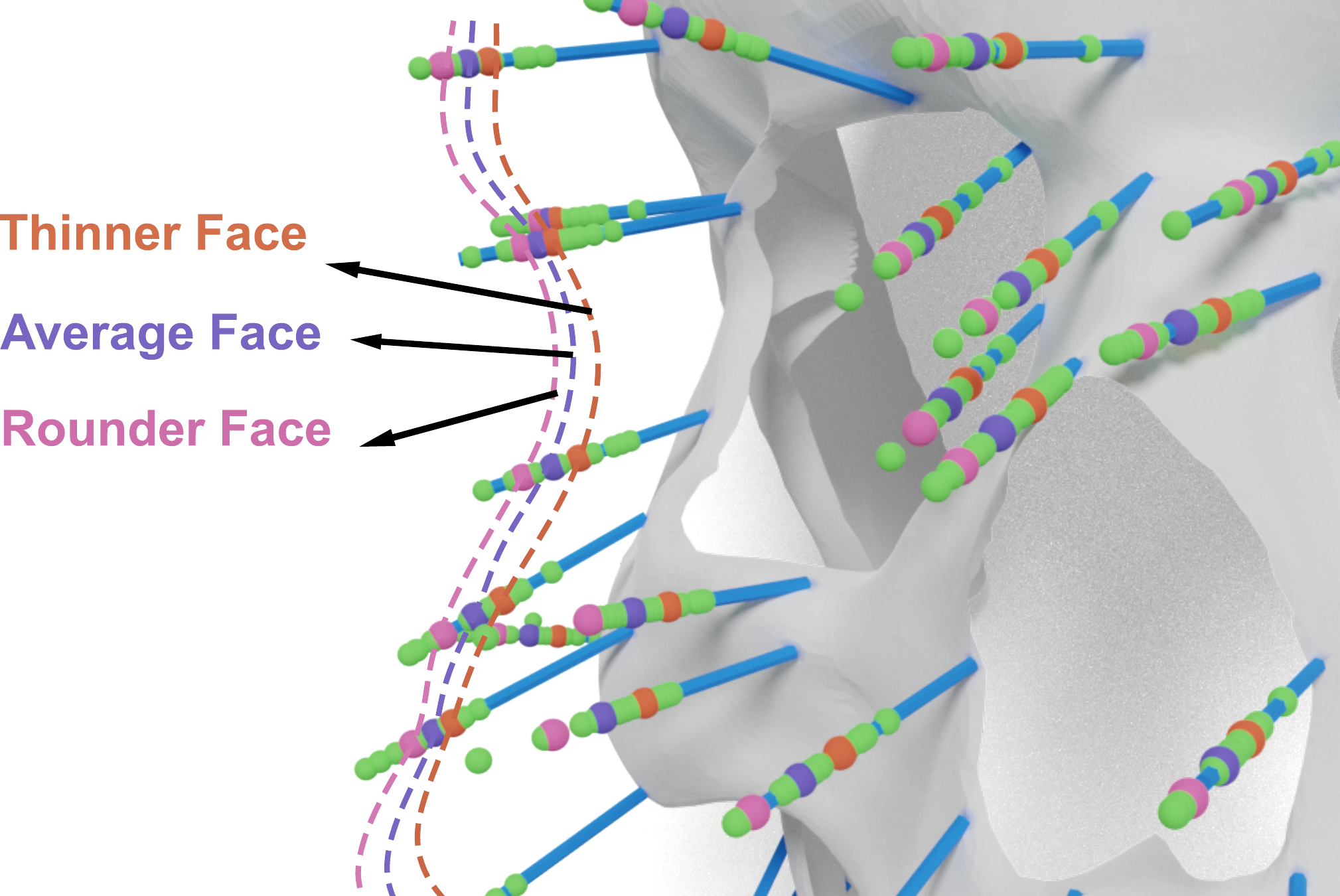} \\
        (b) Adjusting the values of the first principal component\\ provides easy control of tissue depths, and the face shapes \\ change from thin to fat.\\
    \end{tabular}
    \caption{The first principal component can meaningfully provide control on the thickness of tissue depths.
    (a) shows the tissue depth distribution $C$ in the first principal component. 
    Vertical lines represent three representative values, signifying three different combinations of tissue depths, resulting in three types of face shapes: thin, normal, and fat, as illustrated in (b).}
    \label{fig:tissue-depth-pca50}
\end{figure}

We first used principle component analysis (PCA) to project the joint distribution of tissue depths from an $n-$dimension space to a low-dimensional space. 
We observed that the first principal component predominantly governs the tissue depth distribution, accounting for $51.9\%$ of the variance. 
This finding significantly reduces the degree of freedom for the combination of 78 tissue depths.
Consequently, we projected joint tissue depths to the first component axis, denoted as $C$. 
The distribution of $C$ is visualized in Fig.~\ref{fig:tissue-depth-pca50}(a). 
Furthermore, the effect of adjusting the values of the first principal component is illustrated In Fig.~\ref{fig:tissue-depth-pca50}(b), where the directions of the tissues are depicted as blue sticks and the tissue depths in the training set are green spheres.
A more comprehensive discussion of PCA components is included in the \emph{Supplementary}.

When we monotonously sample values from the distribution $C$, the tissue depths change from short to long, and the corresponding face shape changes from thinner to rounder. 
To better visualize the relationship between face shape and the distribution $C$, we select mean values from the top, middle, and bottom $33\%$ of $C$ as representative combinations of tissue depths, and visualize them in Fig.~\ref{fig:tissue-depth-pca50}(b). 
Note that while some related morphable face models also provide correlations between PCA components and face fullness~\cite{Blanz19993DMM}, our model is the first model explicitly guided by tissue depth distribution, directly connecting facial skin to the underlying skull.

\edit{
The above model enables us to modify face shapes based on the joint distributions of all the tissue depths on a skull. And we denote it as the \textbf{TDD-Global} model. 
The TDD-Global model is essential because each value of $C$ represents a valid combination of tissue depths, allowing for easy adjustments to the overall tissue depths based on geometric constraints. 
Specifically, (a) we can use $C=0$ to sample the average tissue depths when we would like to generate average faces with moderate facial thickness.  
(b) We can generate multiple facial reconstructions with different values of $C$ to explore different face shapes and allow the user to select the best fit.
(c) We can select $C$ based on available statistical tissue depth data, such as BMI or fatness level, to align more closely with the specific characteristics of the query skull.
The effectiveness of TDD-Global is validated in face interpolation and face shape modification experiments (see Sec.~\ref{sec:exp-TDD-Global}).
}

\subsubsection{TDD-Regional Model}
\edit{
We integrate the above TDD-Global in our pipeline for automatic facial reconstruction.
The correlation between regions is encoded in the global control model. 
In addition, experts and investigators often require a manual and interactive tool to support local fine-tuning. 
The TDD-Global model alone does not allow for precise modification of specific facial landmarks without impacting other regions. 
On the other hand, modifying single tissue depth alone may result in a non-realistic face shape, because the tissue depths in a local region also adhere to a joint distribution. 
This feature is essential for detailed face editing. 
Hence, we introduced the TDD-Regional Model that offers users additional control for fine-tuning details. 
By enabling localized modification, we intentionally remove interdependencies between regions, and the editing capability can be significantly enhanced.
}

To enable local face shape adjustments, we modify the TDD-Global model to account for the joint distribution of tissue depths in local regions. 
We follow a common facial region partitioning~\cite{marur2014facial, nguyen2022anatomy} (forehead, cheeks, chin, middle region, and mouth) to divide facial landmarks into distinct groups $\mathcal{Y}_l \subset \mathcal{X}$, $k=1,\cdots,5$, as is visualized in Fig.~\ref{fig:face-regions-cheek-nose}(b). 
Then, by employing PCA to model the tissue depth distribution in each group individually, tissue depths can be adjusted locally within each subgroup. 

According to the facial anatomy study~\cite{marur2014facial, nguyen2022anatomy}, facial musculature and neurovascular structures are located throughout faces and can be divided into several regions, each responding to different movements and expressions. 
Our TDD-Regional model can work effectively because it supports fine-tuning local face regions by adjusting regional tissue depths in accordance with relevant anatomical structures. 
\edit{
We provide TDD-Regional as an additional control option for users, alongside our automatic TDD-Global pipeline. 
Our focus is on decomposing tissue depths for visualization purposes, rather than analysis. 
We verified our TDD-Regional model by extensive experiments, as shown in  
Sec.~\ref{sec:exp-TDD-Regional} (\eg, Fig.~\ref{fig:face-regions-cheek-nose}) and the \emph{Supplementary} video/document. 
}

\subsection{Initial Face Generation}
\label{sec:face-generation}

When reconstructing a 3D face from a skull, in addition to geometric constraints, the biological profile derived from the skull can also aid in the reconstruction process. 
The biological attributes often include age, gender, and ancestry, which determine the basic characteristics of the generated face.
\edit{
We provide users with two options to generate an initial face through text prompt depending on how much information of the query skull is available.
By default, when the user doesn't have much information of the skull, they can use the provided template prompt with attributes to synthesize the initial face.
Each attribute has several potential choices for the user to select.
In the other situation, when rich information of the skull is available, the user can add specific descriptions or disable some descriptions in the template to adjust the text prompt to adapt to the biological profile of the skull.
This could generate a more detailed initial face as needed.
}
In this first stage, we generate initial faces following the characteristics of the population group specified by these attributes.

We adopt Stable Diffusion XL~\cite{anonymous2023sdxl} to generate high-resolution 2D portrait images from text prompts. 
We design four main dimensions of facial attributes: gender, ancestry, age, and face shape to generate various faces covering different populations. 
We also add additional prompts to ensure the diffusion model generates appropriate front faces. 
\edit{
In addition to text prompts, users can also use other image generator such as ControlNet~\cite{zhang2023adding} to take edges, scribbles, and sketches as input to synthesize the initial face. 
}

We lift the synthesized portrait images to 3D faces by using DECA~\cite{DECA:feng2021learning}. 
It encodes the image to latent space, consisting of shape, expression, pose, and albedo codes, then decodes the latent code to a 3D face with the FLAME representation.
We transfer the texture and geometry details from the image to the 3D mesh.
\edit{
The face initially generated by Stable Diffusion based on the biological profile serves as a preliminary estimate.
It is then refined using an optimization-based method to align with the skull.
The final 3D face shape is determined by the tissue depth measurements and skull structure, ensuring strict adherence to geometric constraints.
}

\subsection{Anatomy-guided Face Adaptation}
\label{sec:face-editing}

Given an initial 3D face model $F_{in}$, and a set of \emph{facial landmark constraints}, $\mathcal{P}^f=\{ \mathbf{p}_i^f \}$ recommended by our TDD model, we design an anatomy-guided face adaptation model to modify the initial face, ensuring the revised face $F_{out}$ aligns with all facial landmarks. 
Since our face models $\{F\}$ are generated by DECA, they have corresponding latent codes $\{f\}$, and can be decoded into FLAME models $\{G_F\}$; thus, all generated faces share consistent mesh connectivity, $F = G_F(f)$. 
We manually labeled the corresponding $n$ vertex indices on one face as \emph{facial landmarks}, denoted as $\mathcal{Q}^f = \{\mathbf{q}^f_i, i=1,\cdots, n\}$. 
These vertex indices remain consistent across all the generated faces.

Firstly, we estimate a transformation $H$ to make a coarse registration between $\mathcal{Q}^f_{in}$ of initial face $F_{in}$ and $\mathcal{P}^f$.
We solve a Procrustes optimization to estimate the scaling, rotation, and translation parameters of $H$.
With this linear transformation, we transform $\mathcal{P}^f$ to $\mathcal{\Tilde{P}}^f$ as the roughly aligned landmark constraints.
Secondly, we optimize the latent code $f$ by minimizing the objective function $\mathcal{L}$,
\begin{align}
    f_{out}         &= \arg \underset{f}{\min}~~\mathcal{L}(f),\\
    \mathcal{L}     &= \alpha_0\mathcal{L}_{lmk} +\alpha_1\mathcal{L}_{proj} + \alpha_2\mathcal{L}_{sym}.
    \label{eqn:loss}
\end{align}
We design three losses in the objective function and will detail each of them below.

\paragraph{Landmark loss.}
The deviation between the aligned landmark constraints $\mathcal{\Tilde{P}}^f$ and the landmarks $\mathcal{Q}^f$ from the generated face should be minimized: 
\begin{equation}
    \mathcal{L}_{lmk} = \| \mathcal{Q}^f - \mathcal{\Tilde{P}}^f\|_2^2.
\end{equation}

\paragraph{Projection loss.}

\edit{
Due to the limited resolution of the face mesh, the landmarks on the face $\mathcal{Q}^f$ may not perfectly align with the target facial landmarks $\mathcal{\Tilde{P}}^f$. 
To address this, we add a constraint that brings the adapted face mesh as close as possible to the facial landmarks. 
We calculate the shortest distance from each target landmark to the adapted mesh and minimize this distance as a projection loss $\mathcal{L}_{proj}$.
\begin{equation}
    \mathcal{L}_{proj} = \sum_{p\in \mathcal{\Tilde{P}}^{f}} \underset{v\in  F}{\min} \|v - p\|_2^2,
\end{equation}
where $v \in F$ is vertex of the reconstructed face $F$.
}

\paragraph{Symmetry loss.}
Facial symmetry plays an important role in authentic face synthesis that people prefer symmetry faces~\cite{grammer1994human, jones2001facial}.
Hence, we propose a symmetry loss term to encourage the symmetry of the synthesized face.
We first split the facial landmarks $\mathcal{Q}^f$ into three groups, $\{\mathcal{Q}^f_L, \mathcal{Q}^f_M, \mathcal{Q}^f_R\}$, representing on the left, mid, right sides.
The mid-plane $\Pi$ is perpendicular to the line through the eyes, which crosses the tip of the nose and the chin, which is computed by minimizing the least square error of $\mathcal{Q}^f_M$.

For any landmark on the left side $\mathbf{q}^f_i\in \mathcal{Q}^f_L$, we denote $\mathbf{q}^{pf}_i \in \mathcal{Q}^f_R$ as the corresponding landmark on the right side.
We also denote the reflected point across the plane $\Pi$ as  $\mathbf{q}^{rf}_i$.
The symmetry loss function is defined as the Euclidean distance between the actual and expected midpoints $\mathbf{q}^M$, $\mathbf{q}^{M'}$,
\begin{align}
    \mathbf{q}^M = \big(\mathbf{q}^f_i + & \mathbf{q}^{pf}_i\big) / 2,~~~\mathbf{q}^{M'} = (\mathbf{q}^f_i + \mathbf{q}^{rf}_i) / 2,\\
    \mathcal{L}_{sym} &= \sum_{\mathbf{q}^f_i \in \mathcal{Q}^f_L}  \|\mathbf{q}^M - \mathbf{q}^{M'} \|_2^2.
\end{align}

When the face latent code $f$ is optimized, it outputs the geometry of the 3D face $F_{out}$ that conforms to the landmarks and skull constraints.
We transfer the appearance texture and geometry details previously extracted from the 2D portrait to $F_{out}$ as our facial reconstruction.


\begin{figure}[t]
  \centering
  \includegraphics[width=\linewidth]{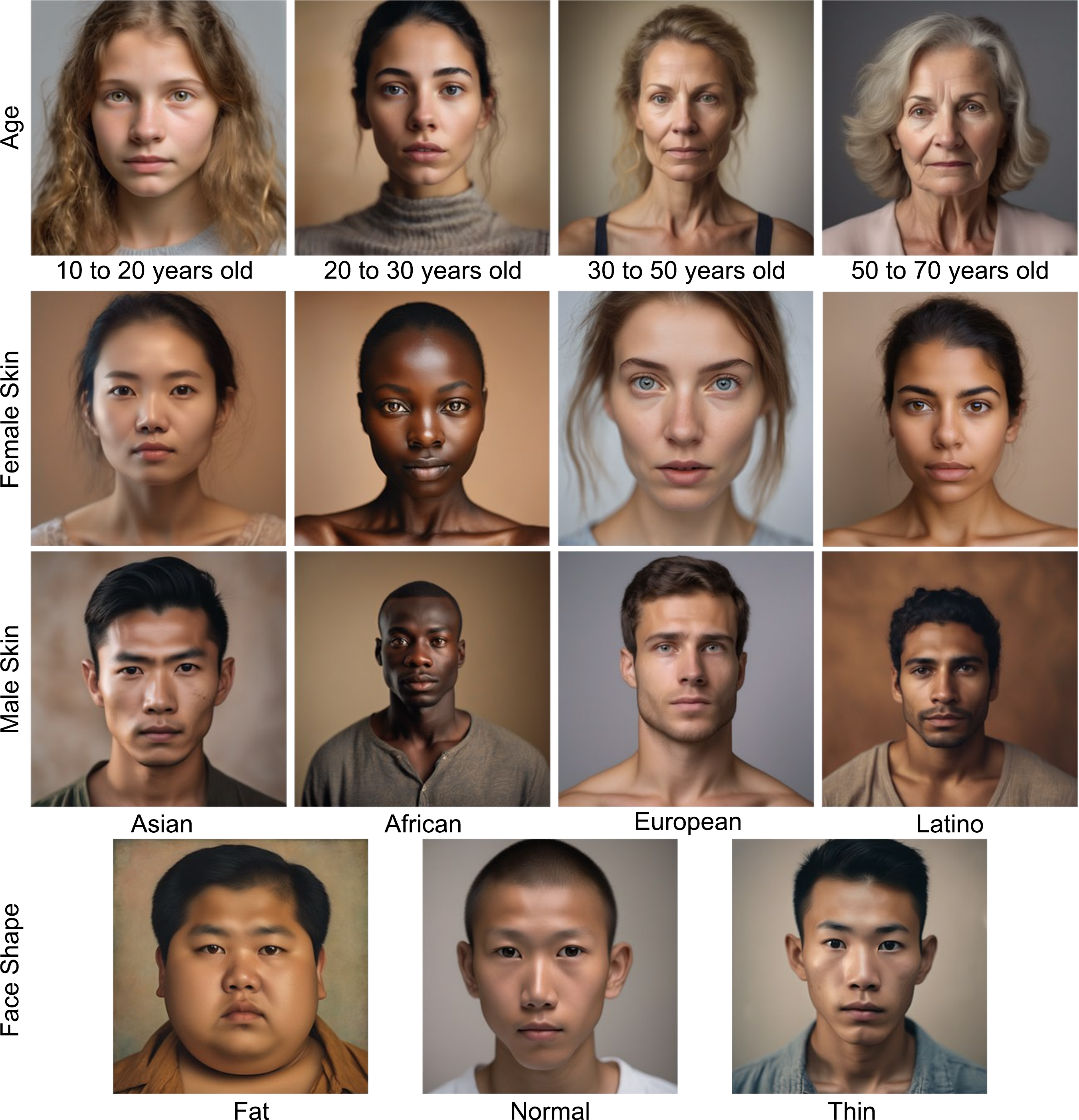}
   \caption{Examples of the synthesized faces by different attributes.}
   \label{fig:sdxl-faces}
\end{figure}

\section{Experiment}

\subsection{Dataset and Error Metric}

We adopted a CT-scan dataset, named \emph{Skull100}, comprising 100 pairs of 3D skulls and faces. These pairs were reconstructed (using \cite{1987A}) from whole-head CT scans. Due to the privacy issue, this dataset doesn't collect face images or appearance data. 
Details of dataset collection are explained in the \emph{supplementary}.
Skull100 is used as the ground truth to evaluate faces reconstructed from their corresponding skulls. 
We divide the dataset into two subsets: 50 pairs for training and 50 for testing. This large test set size is chosen to ensure the statistical significance of our evaluation results. 
We conducted the K-Fold cross-validation to evaluate the distribution bias (Sec.~\ref{sec:cmp-sota}).

We adopted the normalized mean error metric $E$~\cite{NME:kumar2019uglli} to evaluate the geometric deviation of the reconstructed face $F$ and the ground truth $G$, where $F$ and $G$ are represented as triangular meshes.
For each vertex $v$ in the ground truth face $G$, we find the closest vertex $u$ from the reconstructed face $F$ and sum the Euclidean distance as the reconstruction error,
\begin{equation}
    E = \frac{1}{d|G|} \sum_{v \in G}~ \underset{u \in F}{\min}~\|v - u\|_2,
\end{equation}
where $d$ is the distance between the two ears of the ground truth face, and $|G|$ is the total number of vertices in the ground truth face.
Metrics \emph{Mean}, \emph{Max}, and \emph{Std} are the average, maximum, and standard deviation of reconstruction errors $E$ of all test skulls.

\subsection{Initial Face Generation}

Since facial reconstruction should align with the biological profile of the skull, we design the attributes in the text prompts and leverage the Stable Diffusion XL to generate 2D portraits.
We show the examples of the generated 2D face in Fig.~\ref{fig:sdxl-faces}.
The complete text prompt and portrait attributes are reported in \emph{Supplementary}.
We will release this synthesized face dataset with attributes.


\subsection{Evaluation of Tissue Depth Distribution}
\label{sec:exp-tdd}

\subsubsection{TDD-Global Model}
\label{sec:exp-TDD-Global}
The TDD-Global model allows users to use only one parameter $C$ to sample various valid combinations of tissue depths.
The distribution and parameter choice of PCA decomposition are analyzed in the Supp.

\begin{figure*}[t]
    \centering
    \includegraphics[width=0.96\textwidth]{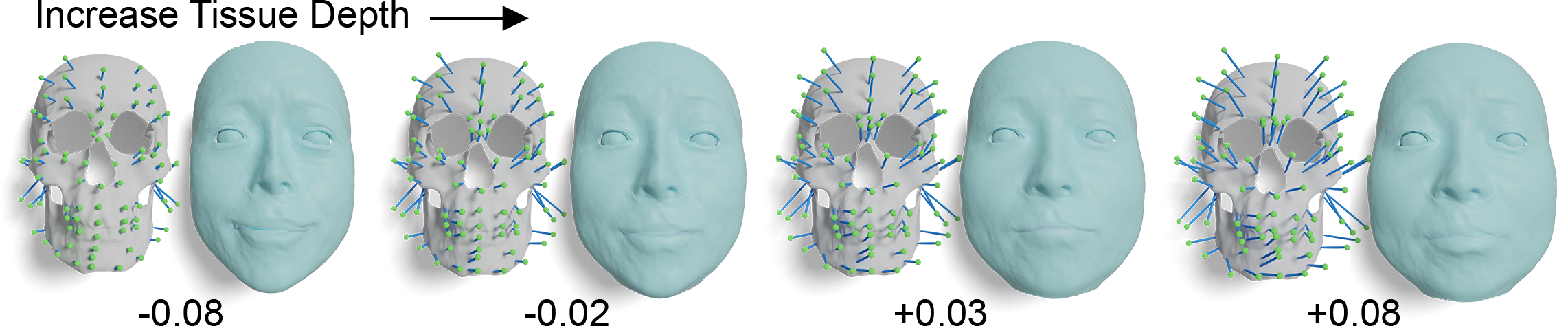}
    \caption{TDD-Global model can explore faces with various shapes. From left to right, we iteratively sampled values of distribution $C$, and the corresponding tissue depths increased from short to long.
    The adapted faces that fit the tissue depth become fatter.}
    \label{fig:pca50-global}
\end{figure*}

In Fig.~\ref{fig:pca50-global}, we enumerated the TDD-Global model to sample different combinations of tissue depths from short to long.
The 3D faces fit these tissue depths well and the face shape becomes fatter from left to right. 
Existing facial reconstruction methods~\cite{HF-GGR:jia2021craniofacial, zhang2022end} only forcibly memorize the distribution of skull and face. 
Instead, our TDD-Global models and discovers the intrinsic pattern of tissue depths and human face shapes.
It allows researchers to efficiently try different tissue depths and examine a variety of facial reconstruction possibilities.

\begin{figure*}[t]
    \centering
    \includegraphics[width=0.96\linewidth]{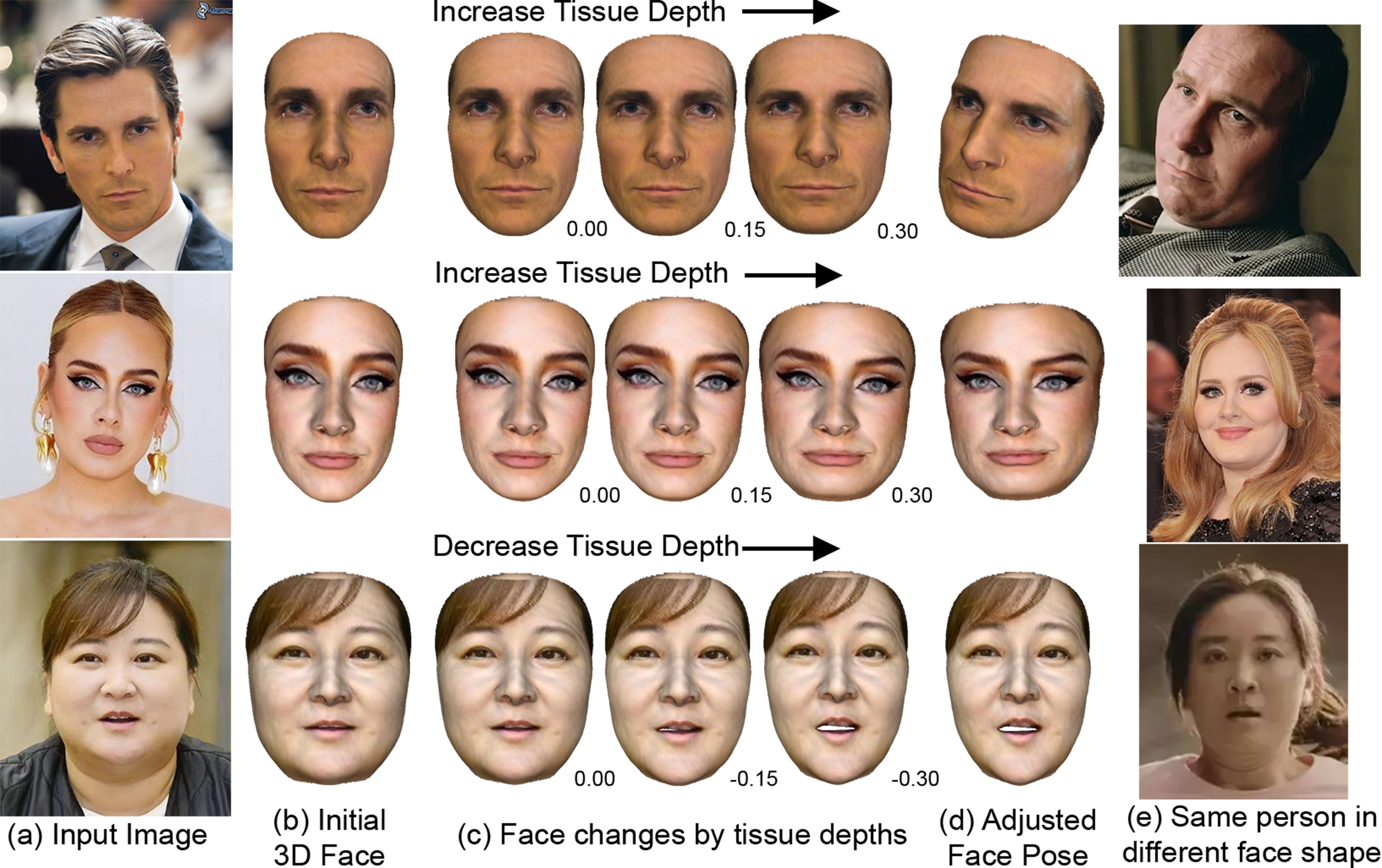}
    \caption{Validation of the face shape modification by TDD-Global model celebrities Christian Bale, Adele Adkins, and Ling Jia. 
    We started from a 2D image (a) and used DECA to create its initial 3D face (b). 
    We estimated proxy skull landmarks and adjusted the tissue depths upon them to tune the face shape to multiple roundness level (c).
    Face pose in (d) was intentionally adjusted for better visual comparisons with (e). 
    (e) show the picture of the same person in a different face shape.}
    \label{fig:face-edit-celebrities}
\end{figure*}

In the second experiment, we modified the face shapes of celebrities to show the ability of 3D face adjustment by the tissue depth model.
Fig.~\ref{fig:face-edit-celebrities} shows the results of three celebrities: Christian Bale, Adele Adkins, and Ling Jia.
\edit{
Using a 2D portrait, the DECA model generates a 3D face of a celebrity, $F_0$, with known facial landmarks $\mathcal{Q}^f$, thanks to DECA's consistent face topology. 
To illustrate the transition of thin to round faces, we sample tissue depths with $C' < 0$ in the TDD-Global model ($TG$) and gradually increase $C'$ to sample different tissue depths $\mathcal{D} = {d_i}$. 
The process is reversed to demonstrate the opposite transition.
We compute the skull landmarks as $\mathcal{P}^s = \mathcal{Q}^f - \mathcal{D}$.
Next, we vary $C'$ to sample different valid tissue depth combinations (e.g., from thin to round faces), where $\mathcal{D}' = TG(C')$, and then calculate the target facial landmarks $\mathcal{P}^f = \mathcal{P}^s + \mathcal{D}'$ based on these skull landmarks.
Finally, the proposed anatomy-guided face adaptation aligns the 3D face $F_0$ to the new facial landmarks $\mathcal{P}^f$, producing the modified face $F'$. We then compare $F'$ with another image of the same person in a different face shape.
}

\begin{figure}[t]
    \centering
    \includegraphics[width=\linewidth]{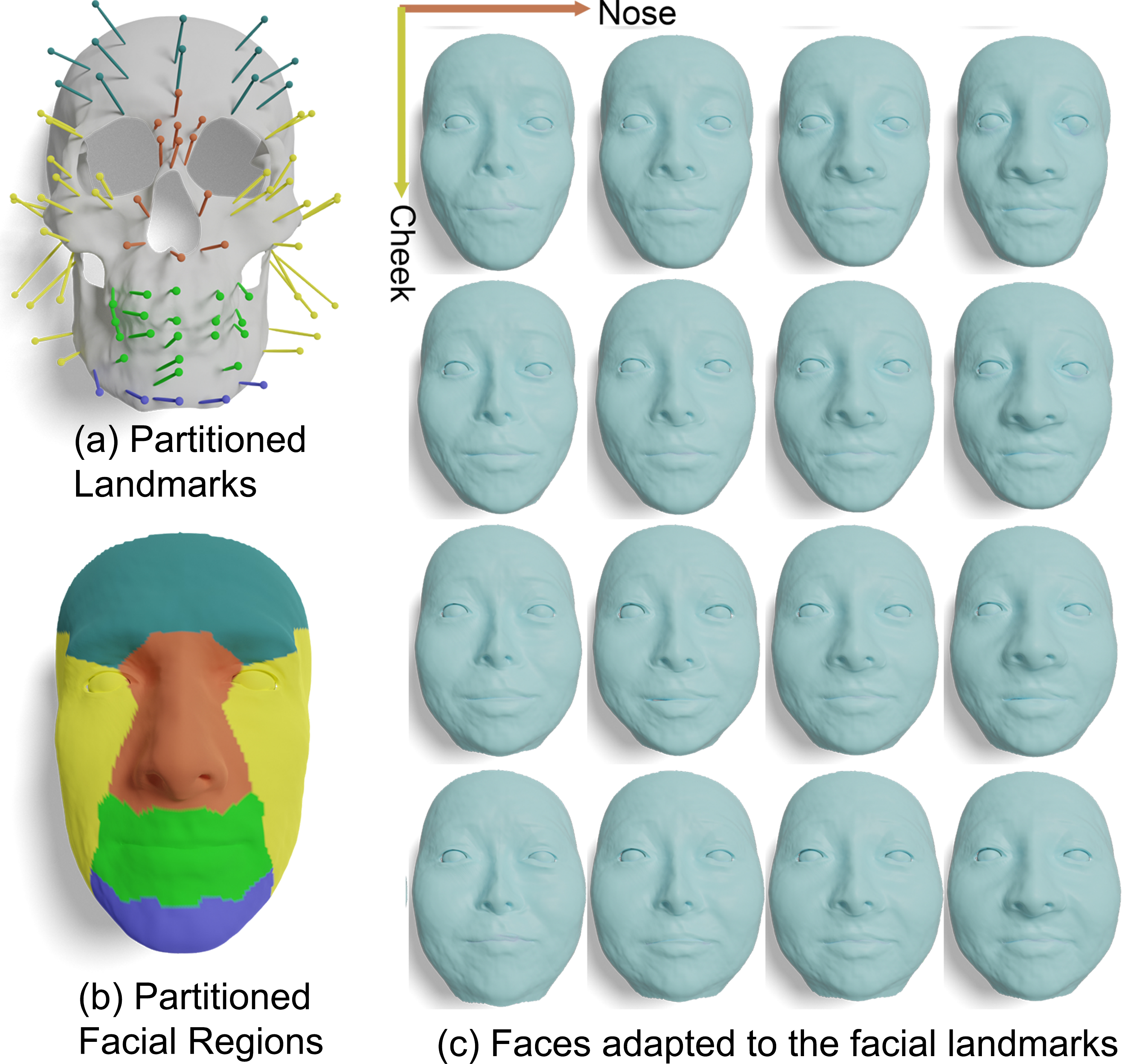}
    \caption{
        Adjusting local face shapes using the TDD-regional model. 
        Tissue depths increase on the nose (from left to right) and cheek (from top to bottom), separately. 
        These modified tissue depths fine-tune facial landmarks, thereby adjusting the face shapes. 
    }
    
    \label{fig:face-regions-cheek-nose}
\end{figure}

\subsubsection{TDD-Regional Model}
\label{sec:exp-TDD-Regional}
We decoupled facial landmarks into distinct regions and applied separate PCA models to model the tissue depth distribution in each region, resulting in the \emph{TDD-Regional} Model. 
This model offers explicit control for manipulating face shapes locally. 
As illustrated in Fig.~\ref{fig:face-regions-cheek-nose}, variations in the landmarks of the nose and cheek regions demonstrate the model's capability. The nose size visibly increases from left to right, while the face shape appears fuller from top to bottom. 
We refer to the \emph{supplementary video} for the animation of face fine-tuning.

\begin{figure*}[t]
  \centering
   \includegraphics[width=0.96\linewidth]{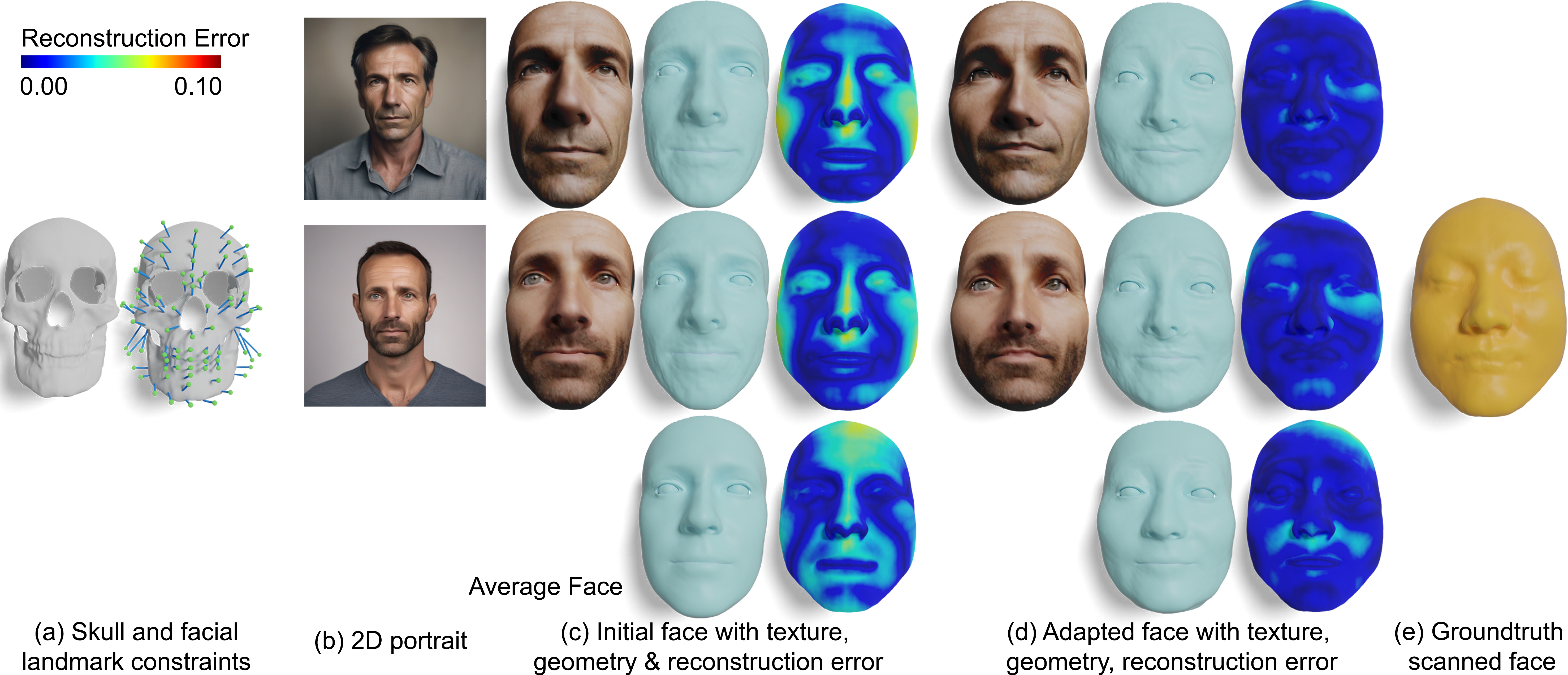}
   \caption{\edit{Validating the influence of initial faces on the final results.
   (a) are the query skulls and facial landmarks.
   (b) is the synthesized portrait image.
   (c) are the initial 3D faces with texture and the reconstruction errors.
   (d) are our facial reconstructions, which and fit the landmark constraints well. 
   The final reconstruction errors are similar: 0.899\% (top), 0.925\% (middle), and 1.19\% (bottom).
   They are close to the GT face (e).
   The bottom row shows the facial reconstruction from the FLAME model's average face. Although our face adaptation module can edit the reconstructed geometry to the GT face, without the proposed initial face generation, using average face lacks geometric details and texture.}
   }
   \label{fig:skull2face-reconstruct1}
\end{figure*}

\begin{figure*}[tb]
    \centering
    \includegraphics[width=\linewidth]{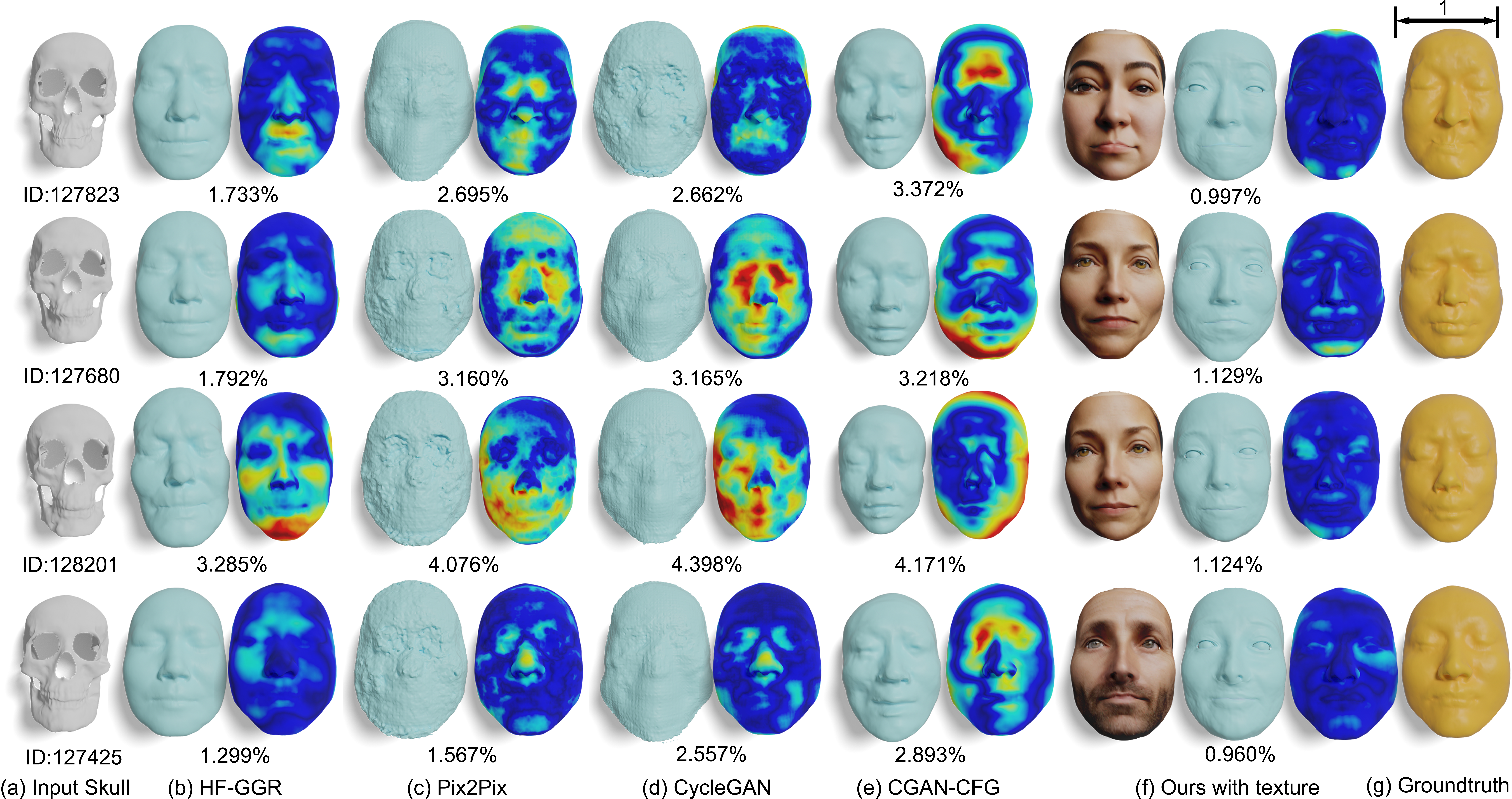}
    \caption{Qualitative comparisons with existing methods. 
    (a) is the input skull.
    (b-e) show the results of HF-GGR~\cite{HF-GGR:jia2021craniofacial}, Pix2Pix~\cite{isola2017pix2pix}, CycleGAN~\cite{CycleGAN2017}, and CGAN-CFG~\cite{zhang2022end}.
    (f) show our reconstructed face with texture and geometry.
    (g) is the CT-scanned GT.
    }
    \label{fig:skull2face-cmp}
\end{figure*}

\begin{figure*}[t]
  \centering
   \includegraphics[width=0.96\linewidth]{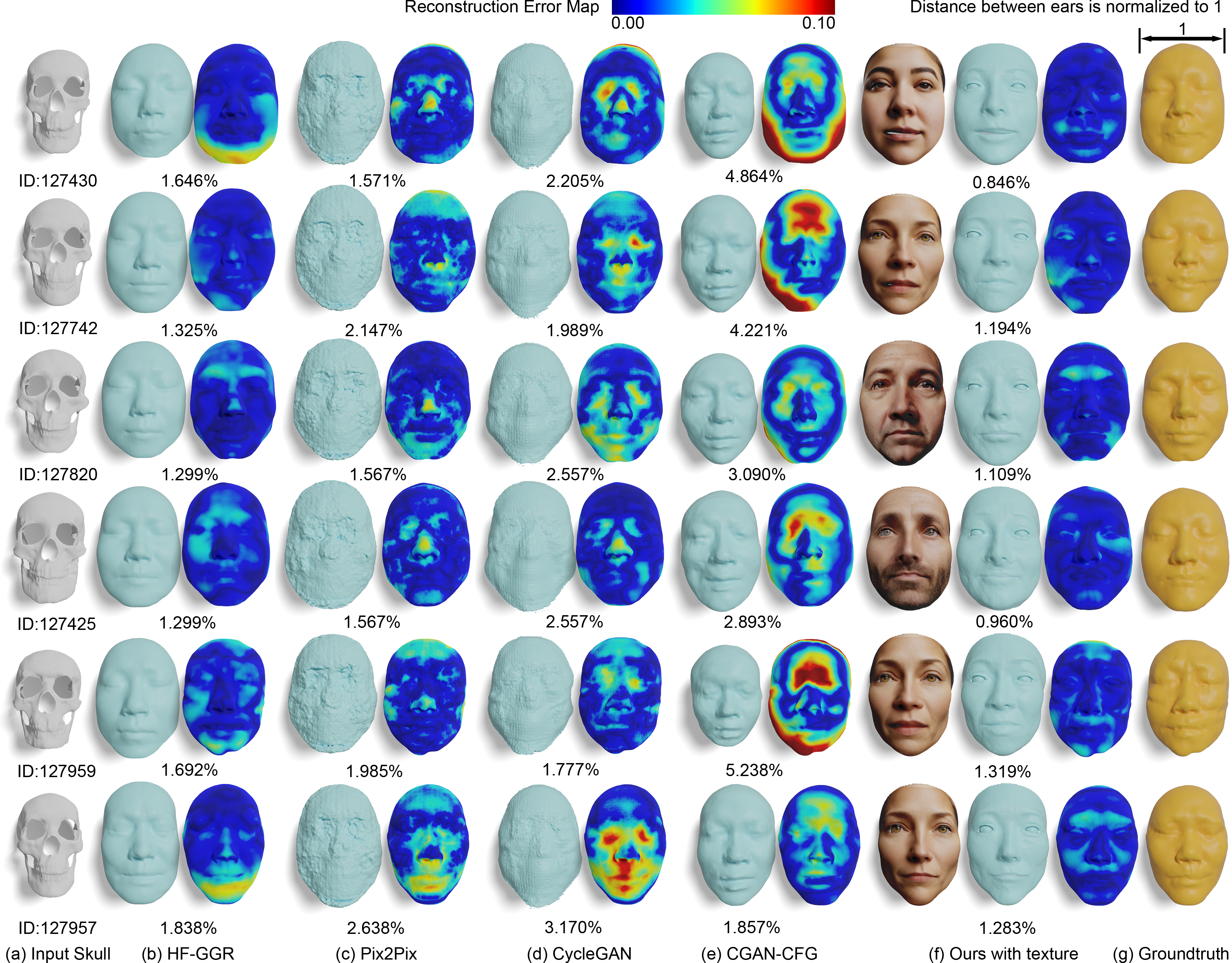}
   \caption{Extensive qualitative comparisons with existing methods. 
    (a) is the input skull.
    (b-e) show the results of HF-GGR~\cite{HF-GGR:jia2021craniofacial}, Pix2Pix~\cite{isola2017pix2pix}, CycleGAN~\cite{CycleGAN2017}, and CGAN-CFG~\cite{zhang2022end}.
    (f) show our reconstructed face with texture and geometry.
    (g) is the CT scanned groundtruth.
   }
   \label{fig:supp-cmp}
\end{figure*}

\begin{figure*}[t]
    \centering
    \includegraphics[width=\linewidth]{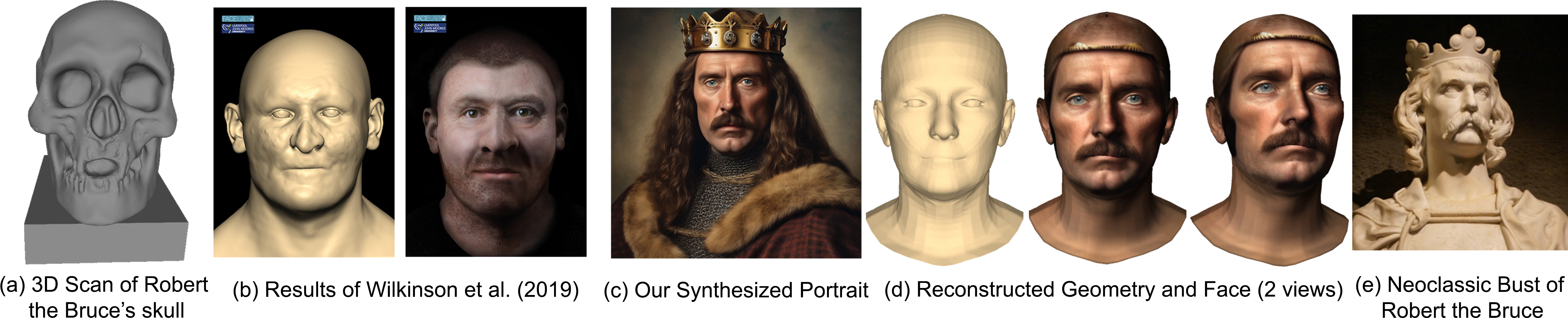}
    \caption{Facial reconstruction demo on historical figure: Robert the Bruce, was King of Scots from 1306 to his death in 1329. (a) is the 3D scan of the skull. (b) show the facial reconstruction by craniofacial experts Wilkinson~\etal~\cite{wilkinson2019search}. (c) is our synthesized portrait. We manually labeled the skull landmarks and used the pretrained TDD-Global to estimate the facial landmarks. (d) show the 3D reconstructed face that fits the facial landmarks. We adjusted the head pose for better visual comparison with reference (e).
    }
    \label{fig:real-robert-bruce}
\end{figure*}


\subsection{Evaluation of Skull-to-Face Reconstruction}

\subsubsection{Influence of Initial Faces on Final Results}
We chose the average tissue depths to compute the extended landmarks as facial constraints.
Fig.~\ref{fig:skull2face-reconstruct1} shows that with face refinement, our face adaptation model can effectively modify the face shape to align with the facial landmarks. 
With facial landmark constraints, the two initial faces, after adaptation, converge to a similar face geometry.
Their geometries still exhibit subtle differences inherited from the initial faces (see error maps in (d)).
\edit{
We also use the FLAME model’s average face as the initial face for adaptation. The bottom row of Fig.~\ref{fig:skull2face-reconstruct1} presents the facial reconstruction results, with final reconstruction errors of 0.899\% (top, face candidate 1), 0.925\% (middle, face candidate 2), and 1.19\% (bottom, average face). Although the reconstructed geometry from the average face is slightly less accurate than the two candidates, it remains close to the ground truth. This demonstrates that our face adaptation module effectively edits the face to fit geometric constraints. Notably, without the proposed initial face generation, the reconstruction from the average face lacks geometric detail and texture.
Our initial face generation serves as a powerful visualization tool, providing users with a textured face from text prompts (several keywords or phrases), which offers valuable reference information.
}

We conducted a quantitative experiment where we randomly synthesized four portraits for each skull and measured the final reconstruction errors.
The reconstruction errors $E$ are similar for different initials: 1.756, 1.673, 1.724, and 1.755.
For each skull, the overall face geometries amongst the final faces (from different initial faces) are similar.
Subtle differences are noticed which are inherited from the initial faces (\eg dimples, wrinkles, and lip thick).

We also conducted a user study to assess the convergence quality of the final facial reconstruction results from different initial faces.
For each question, four initial faces are randomly synthesized.
Three of them are used to reconstruct faces from Skull A, and the other one is used to reconstruct a face from Skull B. 
Users are required to identify the face geometries amongst the reconstructed faces and select the unique face that is reconstructed from Skull B. 

For each question, we randomly synthesized four faces from four different portraits and used the proposed Anatomy-guided Face Adaptation module to adapt to the query skull.
Faces A, B, and C are final reconstructions of one skull. 
Face D is the final reconstruction of the other skull. 
Faces are shuffled in the user study.
Users are asked to compare the reconstructed faces and select the unique one reconstructed from the other skull.
Fig.~\ref{fig:user-study-questions} in supplementary shows two examples of user study questions. 

We collected 500 valid responses from 50 valid users. 
For each question, we count 1 score so the overall score is 10 for 10 questions.
The average correctness is $91.2\%$.
The result of the user study demonstrates that (1) our model can adapt different initial faces and converge to face shapes that are closely related to the query skull, resulting in distinct reconstructions for different skulls.
(2) The reconstructed faces from one skull are not identical, they still preserved subtle geometric details inherited from the initial faces.
Detailed user study result is shown in the supplementary Fig.~\ref{fig:user-study-result}.

The high average correctness of this user study indicates that our model can adapt different initial faces and converge to a face shape space that is related to the query skull, which is distinct from the reconstructed faces that are from the other skulls.

\begin{table}[t]
    \centering
    \caption{Quantitative comparisons of facial reconstructions with existing methods.  The distance between two ears of the groundtruth skin is normalized to 1. 
    \label{tab:skull2face-cmp}}
    \begin{tabular}{@{}lccc@{}}
        \hline
        Method & Mean (\%)$\downarrow$ & Max (\%)$\downarrow$ & Std. (\%)$\downarrow$ \\
        \hline
        HF-GGR~\cite{HF-GGR:jia2021craniofacial} &  2.198 & 4.755 & 0.841\\
        Pix2Pix~\cite{isola2017image} &             2.651 & 4.859 & 0.928\\
        CycleGAN~\cite{zhu2017unpaired} &           2.954 & 4.833 & 0.859\\
        CGAN-CFG~\cite{zhang2022end} &              4.002 & 7.402 & 1.192\\
        \hline
        \textbf{Ours (avg. tissue)} &       \textbf{1.732} &  \textbf{1.775} & \textbf{0.039}\\
        \textbf{Ours (best tissue)} &       \textbf{1.526} &  \textbf{1.571} & \textbf{0.035}\\
        \hline
    \end{tabular}
\end{table}

\subsubsection{Comparisons with SOTA Facial Reconstruction Methods}
\label{sec:cmp-sota}
Quantitative side-by-side comparisons are challenging due to the high cost and privacy concerns associated with joint skull-face CT scans, compounded by the absence of public benchmark data. 
Additionally, the source code for existing methods is not publicly accessible. Despite these limitations, we endeavored to conduct a fair comparison. 
CGAN-CFG~\cite{zhang2022end}, a state-of-the-art DNN-based facial reconstruction method, was tested using our training and testing dataset, with reconstruction results provided by the original authors. 
We also re-implemented CycleGAN~\cite{CycleGAN2017} and Pix2Pix~\cite{isola2017pix2pix} following the guidelines in~\cite{zhang2022end}. 
We chose HF-GGR~\cite{HF-GGR:jia2021craniofacial} as the representative traditional PCA method and also re-implemented their model.

Tab.~\ref{tab:skull2face-cmp} shows the quantitative comparisons of facial reconstructions.
\edit{
We selected three values of $C$—representing thinner, average, and rounder face types—for all test cases and reported both the average and best reconstruction accuracies. 
}
Our method used synthesized portraits as initial faces.
\emph{Ours (avg. tissue)} used the average tissue depths from TDD-Global to compute the facial landmarks and then adapted the initial faces to the final results.
\emph{Ours (best tissue)} utilized three types of tissue depths thin, normal, and fat from TDD-Global as shown in Fig.~\ref{fig:tissue-depth-pca50}.
\edit{
TDD-Regional is provided to allow users to fine-tune the reconstructed face, while TDD-Global was used here for comparison, as it samples a set of reasonable tissue depth values.
}
We selected the smallest reconstruction result as the final face.
Our method outperforms existing methods in all metrics, largely due to the effective adaptation of our loss function to geometric constraints. 
The CT scan resolution is 0.25mm/pixel. The average of the reconstruction error is 3.464 millimeters.

Fig.~\ref{fig:skull2face-cmp} show qualitative comparisons.
CGAN-CFG~\cite{zhang2022end} tends to overfit the training set, leading to worse results on the test set. 
Similarly, Pix2Pix~\cite{isola2017pix2pix} and CycleGAN~\cite{CycleGAN2017} often produce unrecognizable faces as they rely on the availability of large training datasets. 
HF-GGR~\cite{HF-GGR:jia2021craniofacial} is a PCA-based model and typically generates average faces. 
More results are in the \emph{Supplementary}.

Fig.~\ref{fig:supp-cmp} shows extensive comparisons with the existing skull-to-face reconstruction methods.
HF-GGR~\cite{HF-GGR:jia2021craniofacial} is a method that tends to produce average faces that lack details.
CGAN-CFG~\cite{zhang2022end} directly memorizes the relationships between skull and face in the training set, leading to difficulties in generalization to the test set.
Pix2Pix~\cite{isola2017pix2pix} and CycleGAN~\cite{CycleGAN2017} focus on learning image-to-image translation for skull and face images, often resulting in unrecognizable faces due to their requirement of larger training datasets.

To evaluate the distribution bias of training/test set, we used 5-fold cross-validation to split the dataset into training (80\%) and testing (20\%).
The average reconstruction errors on the five testing sets are: 1.761\%, 1.454\%, 1.828\%, 1.575\%, and 1.665\%. The results are consistent with Table~\ref{tab:skull2face-cmp}.

\subsubsection{Validation on Real-world Facial Reconstruction}

We downloaded the 3D-scanned skull from the Anatomical Museums at the University of Edinburgh and used the proposed pipeline to reconstruct the faces for it.
We used the text prompt to generate the initial face according to the historical background.
Fig.~\ref{fig:real-robert-bruce} shows the comparisons between our results and the bust of Robert the Bruce at the National Wallace Monument.
Wilkinson~\etal~\cite{wilkinson2019search} manually sculpted and painted the facial reconstruction. 
Although manual facial reconstruction could be subjective to different researchers~\cite{wilkinson2019search}, the reconstruction of Wilkinson and ours have similar chin, jawline, eyes, and face proportions, which are aligned to the latest depiction~\cite{wilkinson2019search}.

\section{Conclusions}

We proposed a novel anatomy-guided 3D face reconstruction approach that can reconstruct faces from a given skull using both biological profiles of the skull and the statistical tissue depth information defined on physiological landmarks. 
Our pipeline consists of three main components: biological profile-based 3D face initialization, facial landmark prediction using pre-computed tissue depth distributions, 
and facial geometry adaptation to these landmarks. 
These components enable the effective generation of multiple realistic facial variations, encompassing diverse features, skin textures, and adiposity levels. 
Our results demonstrate that the proposed pipeline can reliably generate faces that closely conform to the given skull. 
This work introduces a valuable tool for applications such as forensic investigations and archaeological reconstructions, allowing users to generate and explore potential faces from skeletal remains. 

\emph{Limitations.}
A limitation of our current approach is the constrained complexity of the tissue depth model due to the limited size of the available skull-face dataset. The high cost and prohibitive nature of collecting a large dataset have restricted our ability to train more sophisticated models. 
We will explore potential synthetic data generation techniques to enhance the dataset size and diversity, thereby enabling the exploration of more complex models for facial reconstruction.

\section*{Acknowledgments}
We extend our gratitude to Huijun Han and Zhuowen Shen for their assistance in running experiments, and to Qihua Dong and Xianhe Jiao for their efforts in data collection and cleaning.

\bibliographystyle{IEEEtran}
\bibliography{refs}

\vfill

\clearpage
\setcounter{page}{1}

\appendices

\section{Demo Videos}
In addition to this supplementary material, we refer the reader to the \textbf{supplementary videos} for a more comprehensive explanation and additional results.

\begin{enumerate}
    \item \textbf{demo1-pipeline.mp4} shows the end-to-end pipeline of the facial reconstruction from a skull. This video shows animation to support the first and the second contributions.
    \item \textbf{demo2-landmark.mp4} shows the demo of the third contribution, the animations of 3D facial editing, globally and in local regions.
    \item \textbf{demo3-comparisons.mp4} shows the comparison between our method and existing methods.
    \item \textbf{demo4-king-face.mp4} shows a real-world facial reconstruction example, which uses the skull of Robert the Bruce, King of Scots from 1306 to 1329.
    
\end{enumerate}

\section{Implementation Details}

\subsection{Text Prompt}
\label{sec:text-prompt}
We aim to synthesize 2D portraits that are depicted from the frontal view without occlusion.
Hence, for the automatic face reconstruction pipeline, we designed the following prompt template to synthesize 2D portraits as initial faces,

\begin{quote}
    \emph{prompt}: A photo of a human face, facing the camera, full head visible, sharp focus, fine art portrait photography, with \{Age\}, \{Ancestry\}, \{Gender\}, and \{Face Shape\}.
\end{quote}

The attributes $\{~\}$ are replaced by the values in Table~\ref{tab:sdxl-attributes}. 
We also used the negative prompt to ensure no accessories such as hats and glasses occluding the face,
\begin{quote}
    \emph{negative prompt}: Hat, glasses, black and white, cartoon, anime.
\end{quote}

To reconstruct the face of Robert the Bruce, King of Scots, from the scanned skull, we  searched for his related information~\cite{barrow2013robert}, and composed the following text prompt to depict his appearance and infer his initial face: 
\begin{quote} 
\emph{prompt}: A photo of a King of Scots in 14 century. A powerful and strong Scottish male. He has a large head dark brown hair and brown eyes. He also has a muscular neck and strong frame, preparing him for the brutality of medieval warfare. A crown and royal attire. Looking at the camera. Realistic portrait.
\end{quote}

\subsection{Network Weights and Hyper-parameters}
We used PCA to fit the tissue depths of the training set. 
To provide local control, we further divided the face into five regions and used PCA to fit tissue depths in each region.
We kept the principal component with the largest eigenvalue for both global control and local control.
We used the pre-trained models and fixed their weights in our pipeline (Stable Diffusion XL, DECA encoder, and FLAME decoder).

In Eqn.~\ref{eqn:loss}, we set $\alpha_0 = 5$, $\alpha_1=1$, $\alpha_2 = 1$.
We use AdamW~\cite{Adam:loshchilov2017decoupled} to optimize the face latent code $f$ with an initial learning rate of $10^{-2}$. 
The learning rate is multiplied by $0.2$ for every $200$ iterations, with $1000$ iterations in total.

\subsection{Dataset Processing}
The whole-head CT scans were obtained by using a clinical multi-slice CT scanner system which provides CT images with a resolution of $512 \times 512$ (pixel size $0.5 \times 0.5mm$, inter-slice thickness $0.75mm$). 
After extracting meshes from the CT data, the heads and their underlying skulls are normalized into a unified Frankfurt coordinate system~\cite{Deng11FSI} to eliminate the inconsistency in posture and scale.

\subsection{Pipeline Running Time}

We calculated the overall running time on a single NVIDIA 4090 GPU.
The 2D portrait is synthesized by a pre-trained Stable Diffusion XL model (7s).
The initial 3D face is reconstructed from the synthesized 2D image by a pre-trained DECA model (2s).
Then we adapt the face to match the landmark constraints and use a back-propagation to optimize the face latent code (5s).
The overall time consumption is around 14 seconds for facial reconstruction from a skull.

\begin{table}[tb]
  \centering
  \caption{Attributes used for initial face generation to match the biological profile of the skull.}
  \label{tab:sdxl-attributes}
      \begin{tabular}{@{}ll@{}}
        \hline
        Attributes & Values \\
        \hline
        Age & 10 to 20, 20 to 30, 30 to 50, 50 to 70\\
        Ancestry & African, Asian, European, Latino\\
        Gender & Female, Male\\
        Face Shape & Normal, Fat, Thin\\
        \hline
      \end{tabular}

\end{table}
  

\begin{figure}[t]
    \centering
    \includegraphics[width=\linewidth]{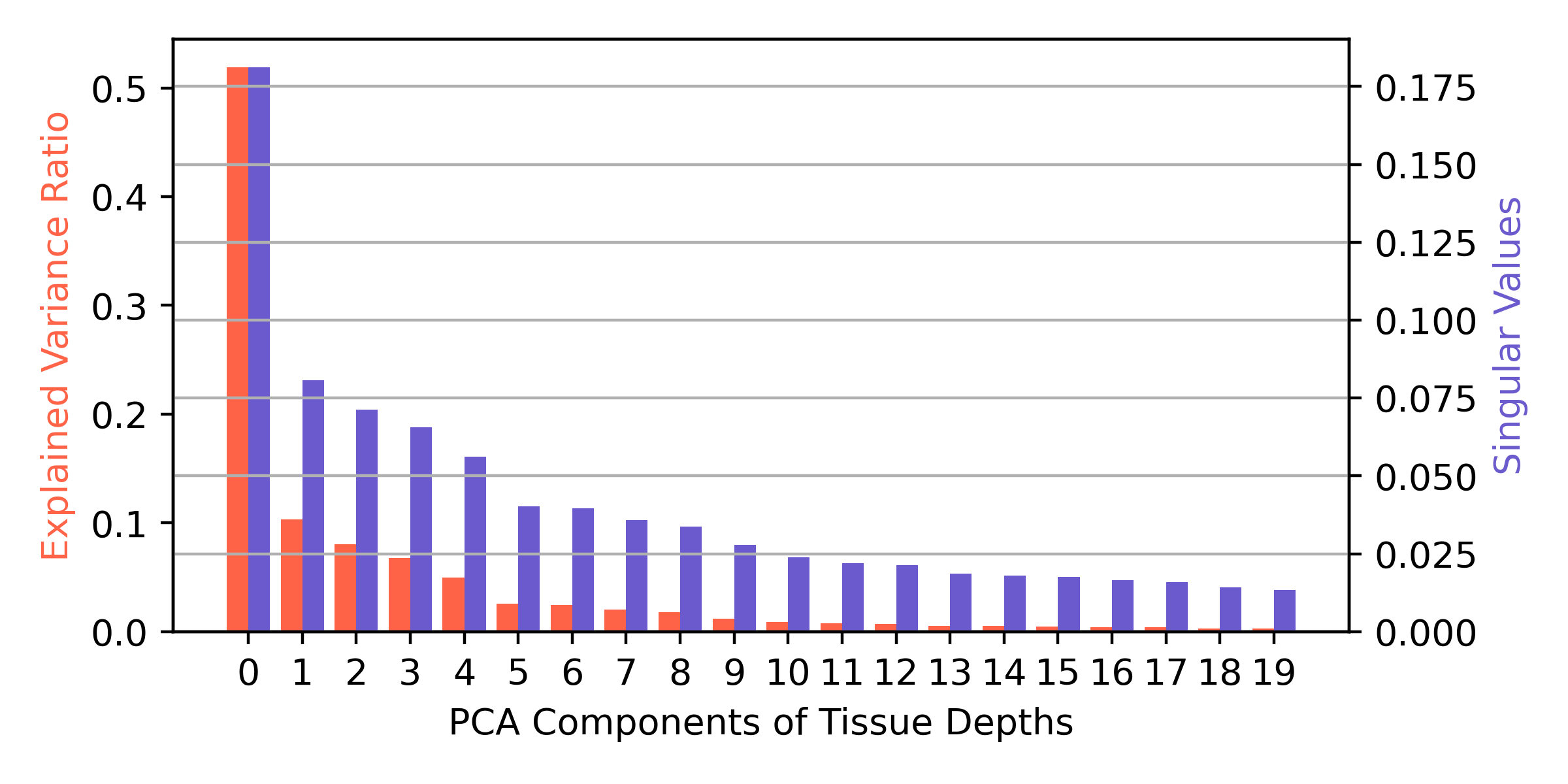}
    \caption{
    PCA decomposition of tissue depths in the training set, with the 1st component dominating the distribution with $51.9\%$ of the variance.
    }
    \label{fig:tissue-depth-pca50-var}
\end{figure}

\section{Ablation Study on Face Adaptation}

We conducted an ablation study to evaluate the impact of various loss functions via three settings: 
(1) using only $\mathcal{L}_{lmk}$,  
(2) using $\mathcal{L}_{lmk}$ + $\mathcal{L}_{proj}$, 
and (3) our complete loss. 
Tab.~\ref{tab:adaptation-loss} reports the average reconstruction errors.
We computed the errors in the physical metric (millimeter) by the average face width (20cm).
The projection loss $\mathcal{L}_{proj}$ aids in reducing the distance between the reconstructed face and landmark constraints. 
The symmetric loss $\mathcal{L}_{sym}$ enhances the face realism by promoting symmetry. 
The second and third rows show the face adaptation results of the same skull, using two different initial faces.

\begin{figure}[t]
    \centering
        \includegraphics[width=\linewidth]{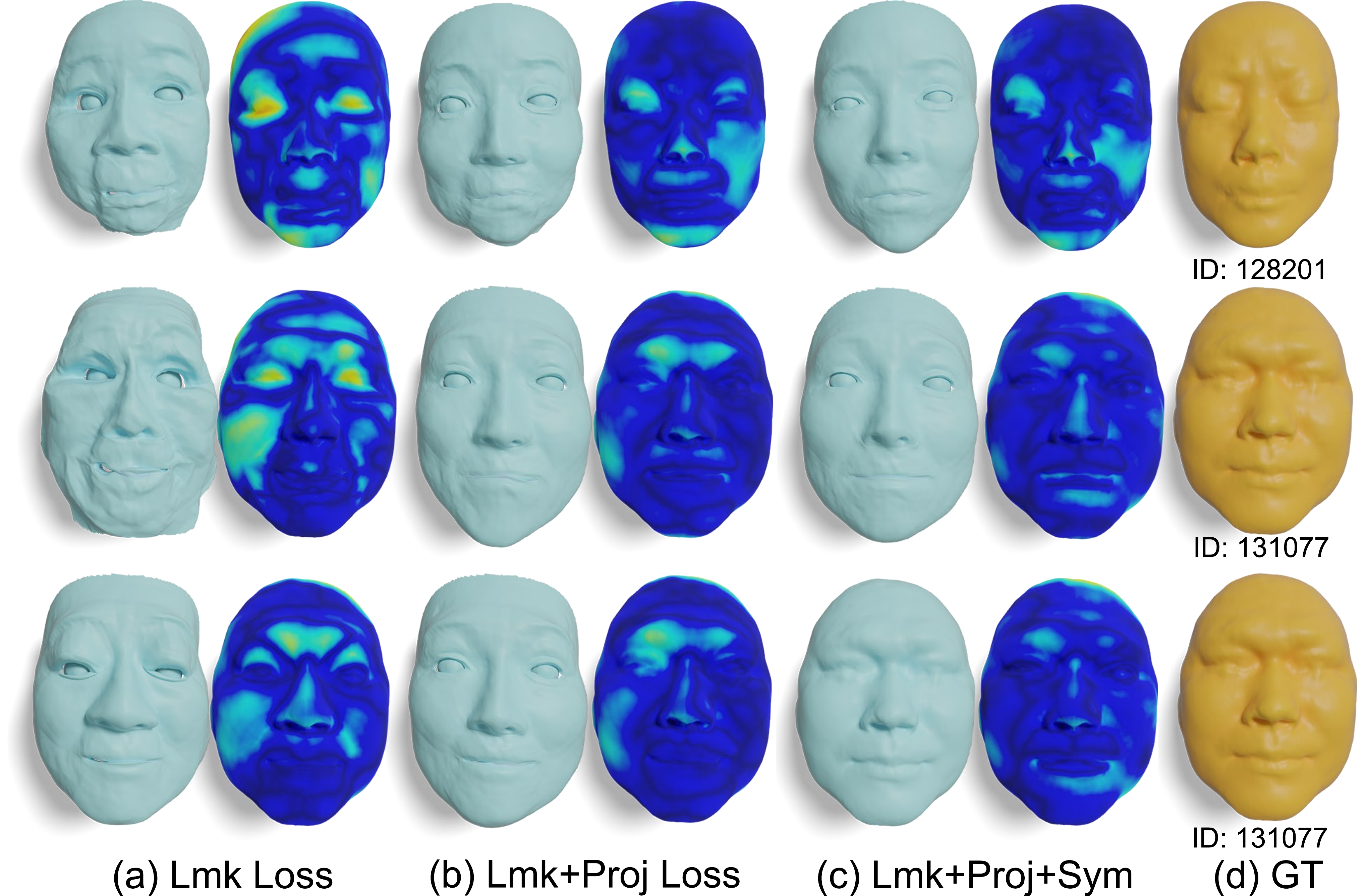}
        \caption{Ablation study on loss functions. Projection loss $\mathcal{L}_{proj}$ helps reduce face deviation from landmark constraints; symmetric loss $\mathcal{L}_{sym}$ enhances face realism.}
        \label{fig:ablation-loss}
\end{figure}

Fig.~\ref{fig:ablation-loss} shows the reconstruction results using different loss functions. 
The proposed projection loss $\mathcal{L}_{proj}$ and symmetric loss $\mathcal{L}_{sym}$ help enhance the face realism.
A quantitative ablation study is reported in the main text.

\begin{table}[h]
    \centering
    \caption{Reconstruction errors before vs after adaptation and loss functions. }
    \label{tab:adaptation-loss}
    \begin{tabular}{@{}lcc@{}}
        \hline
        \multirow{2}{*}{Ablation Study} & Mean Recon. & Mean Recon.  \\
         & (\%) & (millimeter) \\
         \hline 
         1. Before adaptation (from DECA)         & 2.631 & 5.262 \\
         2. w/ $\mathcal{L}_{proj}$               & 2.273 & 4.547 \\
         3. w/ $\mathcal{L}_{lmk}$                & 1.836 & 3.672 \\
         4. w/ $\mathcal{L}_{lmk} + \mathcal{L}_{proj}$ & 1.758 & 3.576 \\
         5. After adaptation  & \multirow{2}{*}{\textbf{1.732}} & \multirow{2}{*}{\textbf{3.464}}\\
         ($\mathcal{L}_{lmk} + \mathcal{L}_{proj} + \mathcal{L}_{sym}$) & & \\
        \hline
    \end{tabular}
\end{table}

\section{User Study}

\begin{figure}[tb]
    \centering
        \includegraphics[width=\linewidth]{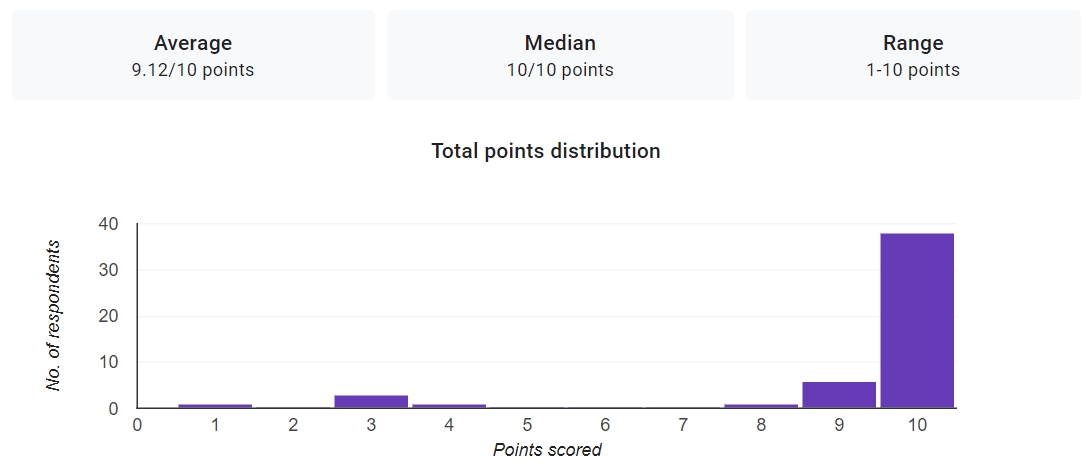}
        \caption{User study of convergence quality of final facial reconstruction from different initial faces. We collected 500 valid responses from 50 valid users. The result shows that our model can adapt different initial faces and converge to a face shape space that is related to the query skull, which is distinct from the reconstructed faces that are from the other skulls. }
        \label{fig:user-study-result}
\end{figure}

\begin{figure}[t]
    \centering
        \includegraphics[width=\linewidth]{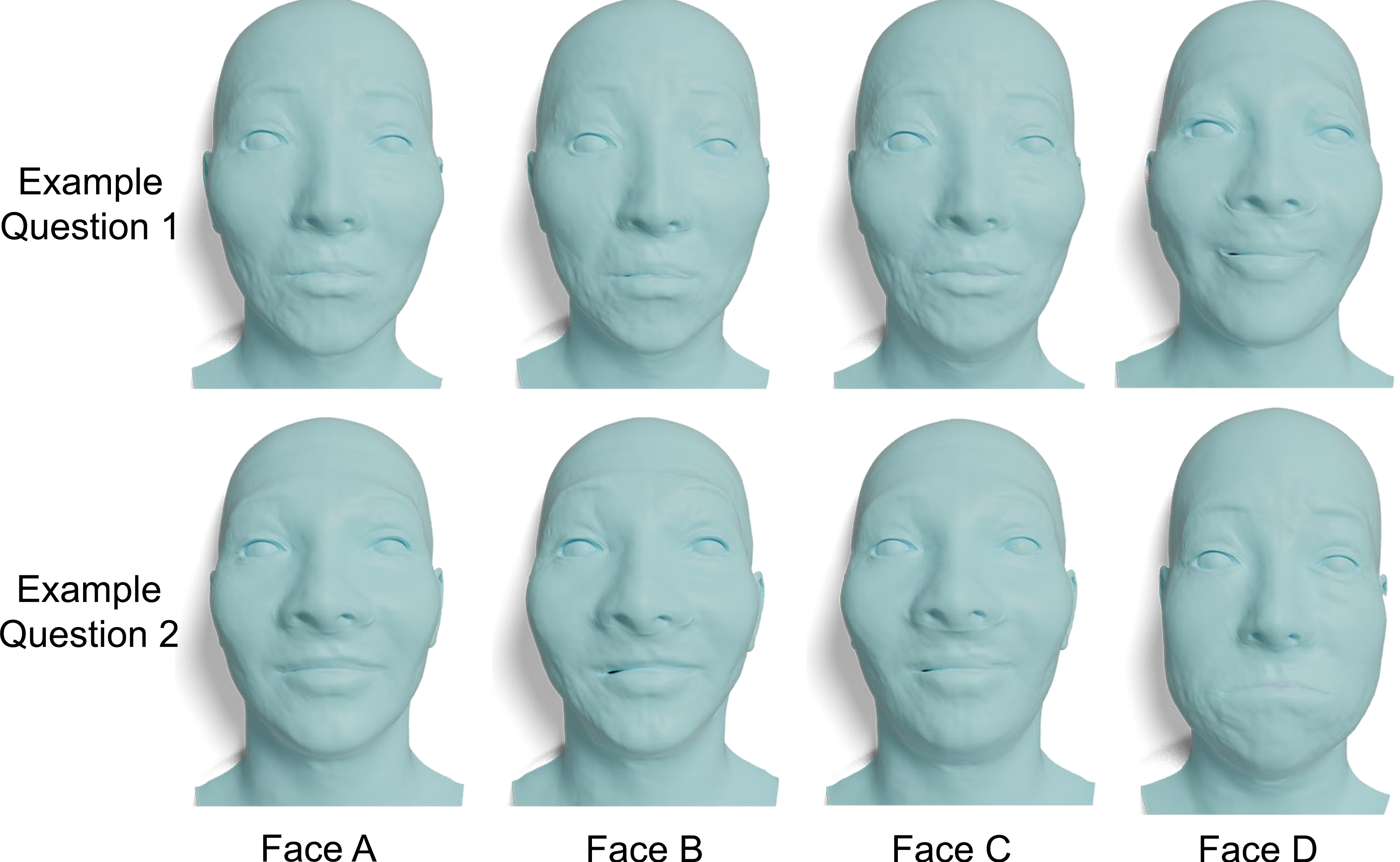}
        \caption{Example questions in the user study. Faces A, B, and C were reconstructed from one skull, while Face D was reconstructed from the other skull. Faces are shuffled in the user study.}
        \label{fig:user-study-questions}
\end{figure}

Fig.~\ref{fig:user-study-result} shows the detailed user study result.
Fig.~\ref{fig:user-study-questions} shows two examples of user study questions.

\section{Analysis on Tissue Depth Distribution}

We used PCA to decompose the joint distribution of tissue depths from the training set of the skull-skin dataset.
We observed that the first component dominates the variances of PCA results, as Fig.~\ref{fig:tissue-depth-pca50-var} shown.

\begin{figure*}
    \centering
    \begin{tabular}{cc}
        \includegraphics[width=0.41\linewidth]{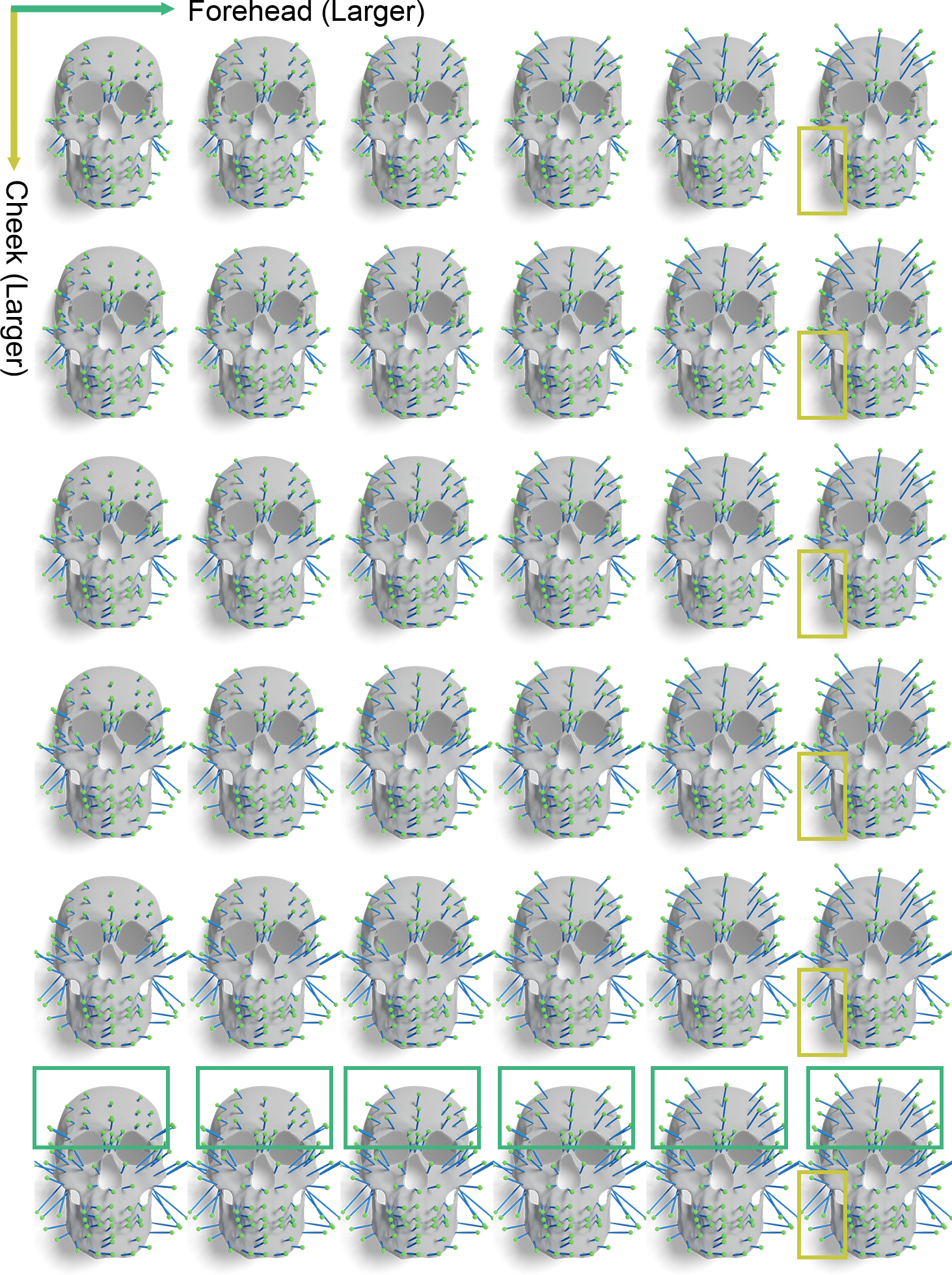} &
        \includegraphics[width=0.41\linewidth]{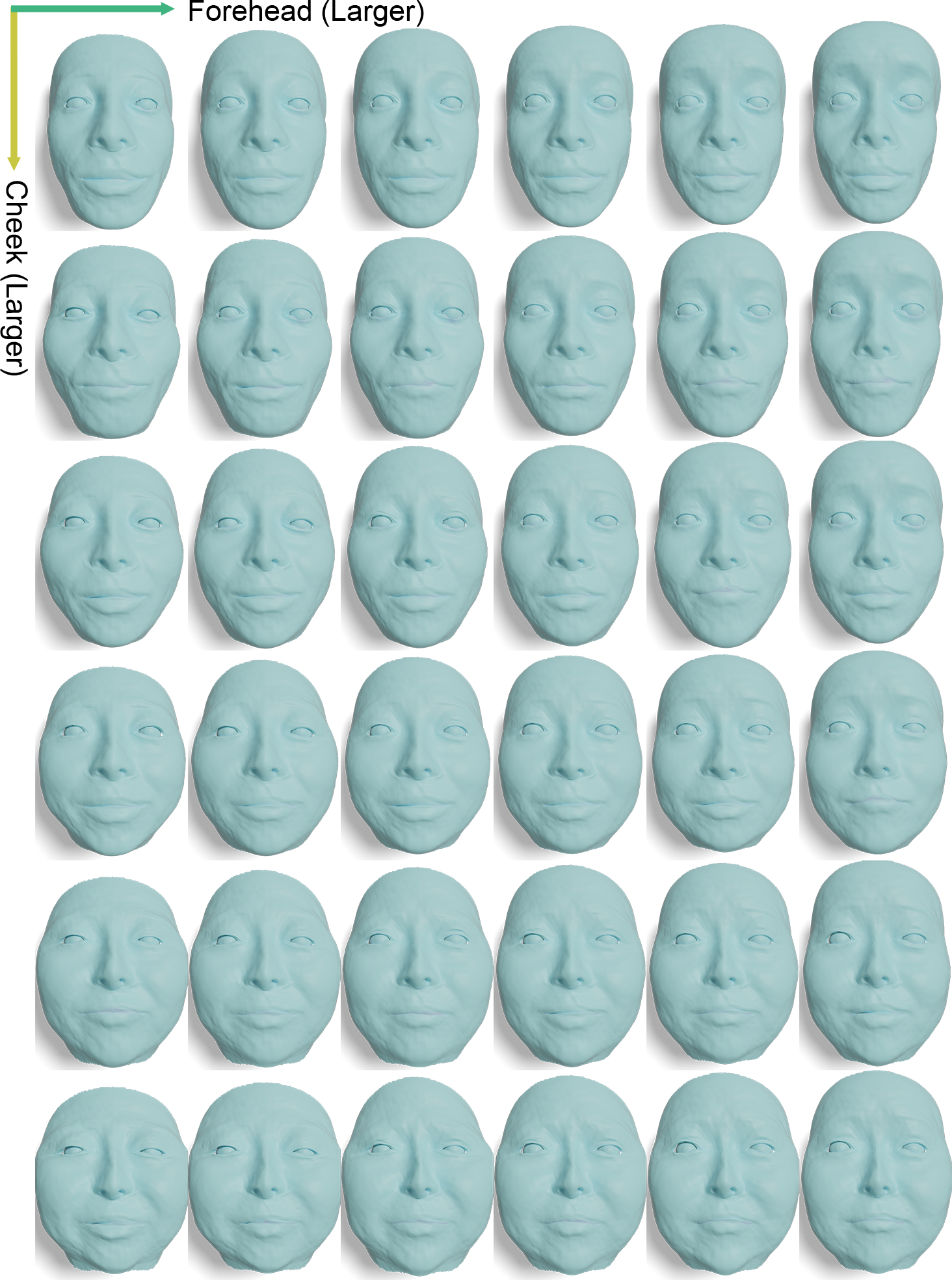}\\
        (a) Changes of Tissue Depths & (b) Faces Adapted to the Facial Landmarks
    \end{tabular}
    \caption{
        Adjusting local face regions by TDD-Regional. 
        Tissue depths increase on forehead (from left to right) cheek (from top to bottom). The resulting faces closely align with the corresponding landmarks.
        Rectangles highlight the affected regions.
    }
    \label{fig:supp-face-regions-cheek-forehead}
\end{figure*}

\begin{figure*}
    \centering
    \begin{tabular}{cc}
        \includegraphics[width=0.41\linewidth]{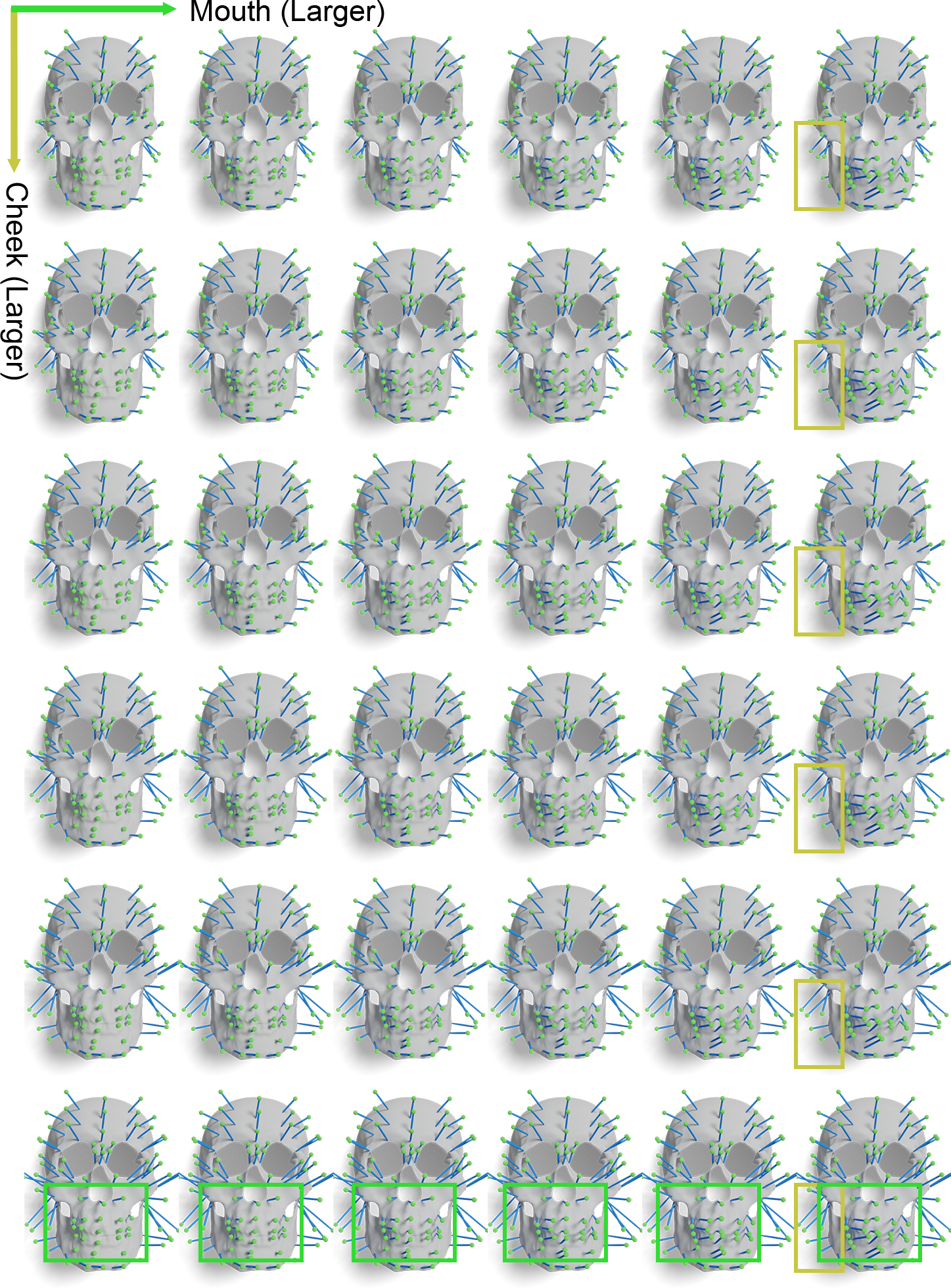} &
        \includegraphics[width=0.41\linewidth]{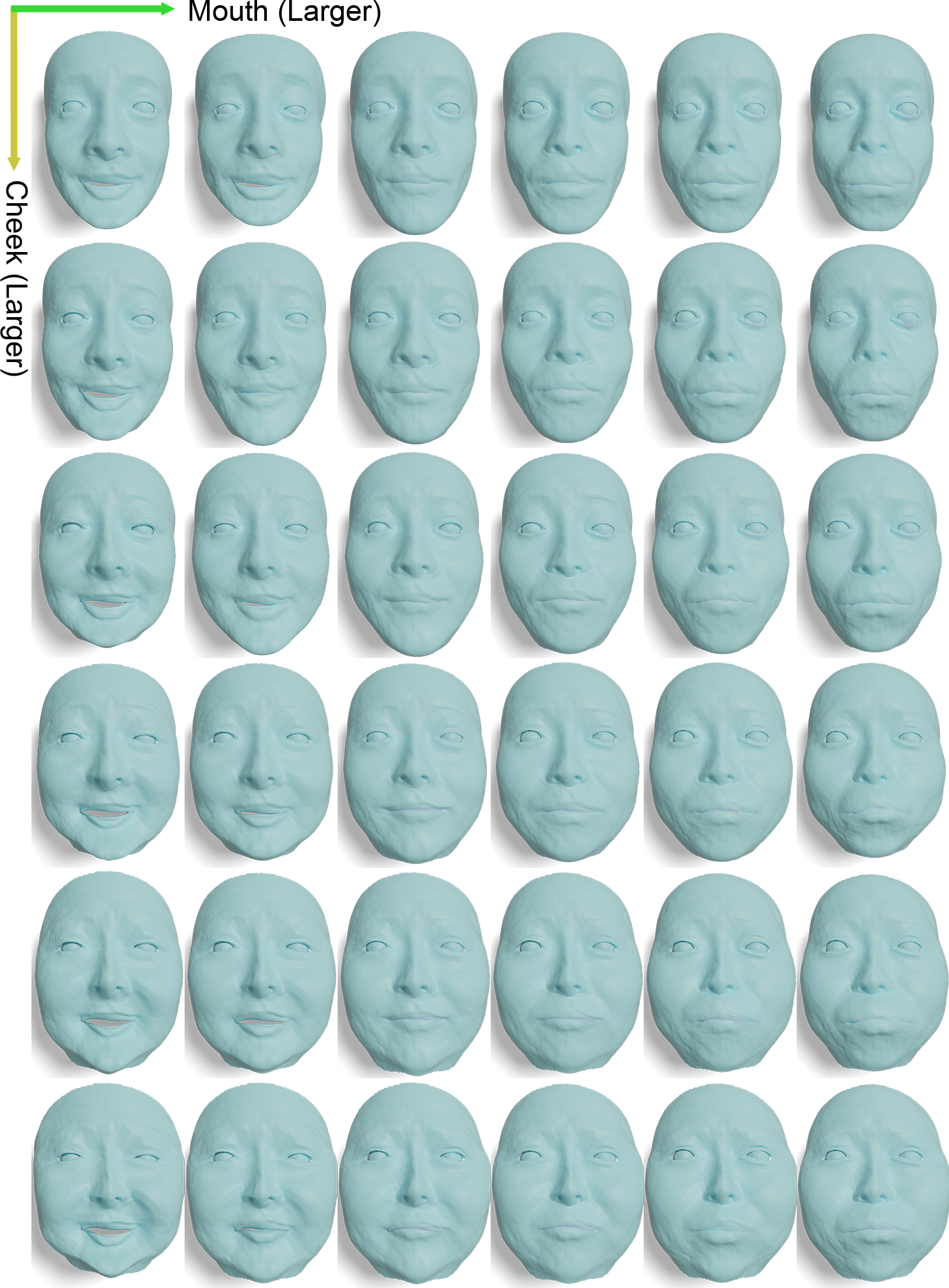}\\
        (a) Changes of Tissue Depths & (b) Faces Adapted to the Facial Landmarks
    \end{tabular}
    \caption{
        Adjusting local face regions by TDD-Regional.
        Tissue depths increase on the mouth (from left to right) and cheek (from top to bottom). The resulting faces closely align with the corresponding landmarks.
        Rectangles highlight the affected regions.
    }
    \label{fig:supp-face-regions-cheek-mouth}
\end{figure*}

\section{Detailed Results of Regional Control}

We project the tissue depths in each face region to the first component of the PCA decomposition.
Then users can intuitively and conveniently control the face shape by tuning one parameter.
We show several examples below to demonstrate how users can intuitively and conveniently control the face shapes. For each example, we randomly select two different regions to visualize the results when tuning multiple regions at the same time.

\paragraph{Forehead and cheek control.}
Fig.~\ref{fig:supp-face-regions-cheek-forehead} shows the adjustment of the forehead and cheek regions by using different combinations of tissue depths.
Our face adaption module can consistently fit the face to the landmark constraints well.

\paragraph{Mouth and cheek control.}
We further show the adjustment of the mouth and cheek regions by using different combinations of tissue depths in
Fig.~\ref{fig:supp-face-regions-cheek-mouth}.

\end{document}